\documentclass[letterpaper, 10 pt, conference]{ieeeconf}  % Comment this line out if you need a4paper

\IEEEoverridecommandlockouts    % This command is only needed if 
\usepackage{graphics} % for pdf, bitmapped graphics files
\usepackage{amsmath} % assumes amsmath package installed
\usepackage{amssymb}  % assumes amsmath package installed
\usepackage{graphicx} % for pdf, bitmapped graphics files
\usepackage{multirow} 
\usepackage[labelformat=simple]{subcaption}
\usepackage{xcolor}
\usepackage{mathtools}
\usepackage[ampersand]{easylist}
\usepackage{boldline}
\usepackage{booktabs}
\usepackage{adjustbox}
\usepackage{mathtools}
\usepackage{algorithm}
\usepackage[noend]{algpseudocode}
\usepackage{hyperref}
\usepackage{xspace}

\newcommand{\reffig}[1]{Fig.~\ref{#1}}
\newcommand{\reftab}[1]{Table~\ref{#1}}
\newcommand{\refsec}[1]{Section~\ref{#1}}
\newcommand{\etal}{\textit{et al.}\xspace}

\newcommand\tableheader[2]{%
  \multicolumn{1}{c}{\parbox{#1}{\centering #2}}
}

\DeclarePairedDelimiter{\norm}{\lVert}{\rVert}

\renewcommand{\deg}{^{\circ}}
\renewcommand{\vec}[1]{\mathbf{#1}}

\algnewcommand\algorithmicforeach{\textbf{for each}}
\algdef{S}[FOR]{ForEach}[1]{\algorithmicforeach\ #1\ \algorithmicdo}

\renewcommand\algorithmicdo{}
\makeatletter
\renewcommand{\ALG@beginalgorithmic}{\footnotesize}
\makeatother

\newlength{\tempdima}
\newcommand{\rowname}[1]% #1 = text
{\rotatebox{90}{\makebox[\tempdima][c]{#1}}}

\title{\LARGE \bf
  Sparse 3D Topological Graphs for Micro-Aerial Vehicle Planning
}

\author{Helen Oleynikova, Zachary Taylor, Roland Siegwart, and Juan Nieto\\
Autonomous Systems Lab, ETH Z{\"u}rich%
}
\begin{document}

\maketitle
\thispagestyle{empty}
\pagestyle{empty}

%%%%%%%%%%%%%%%%%%%%%%%%%%%%%%%%%%%%%%%%%%%%%%%%%%%%%%%%%%%%%%%%%%%%%%%%%%%%%%%%
\begin{abstract}
Micro-Aerial Vehicles (MAVs) have the advantage of moving freely in 3D space. However, creating compact and sparse map representations that can be efficiently used for planning for such robots is still an open problem. In this paper, we take maps built from noisy sensor data and construct a sparse graph containing topological information that can be used for 3D planning. We use a Euclidean Signed Distance Field, extract a 3D Generalized Voronoi Diagram (GVD), and obtain a thin skeleton diagram representing the topological structure of the environment. We then convert this skeleton diagram into a sparse graph, which we show is resistant to noise and changes in resolution. We demonstrate global planning over this graph, and the orders of magnitude speed-up it offers over other common planning methods. We validate our planning algorithm in real maps built onboard an MAV, using RGB-D sensing.
\end{abstract}

%%%%%%%%%%%%%%%%%%%%%%%%%%%%%%%%%%%%%%%%%%%%%%%%%%%%%%%%%%%%%%%%%%%%%%%%%%%%%%%%
\section{Introduction}
One of the most fundamental problems in robotics is planning paths through known maps, often referred to as global planning.
While there are multiple classes of solutions to this problem, including sampling-based methods like RRTs~\cite{karaman2011sampling} and search-based methods like A*~\cite{hart1968formal}, one class of solutions that has not been used for many real problems in 3D: topological planning.

Extracting the topology of a 2D map built by a ground robot to use for planning has been a well-studied topic, including methods that allow the topological graph to be created incrementally and maintained online~\cite{thrun1998learning,lau2013efficient,kalra2009incremental}.
Most of these methods start with a Euclidean Signed Distance Field (ESDF), also known as Euclidean Distance Transform (EDT), of the space, where each point in a 2D grid stores its distance to the nearest obstacle.
From the ESDF, it is then possible to generate the Generalized Voronoi Diagram (GVD) of this space by finding all the points that are equidistant from two or more obstacles. This set of points represents the `ridges' in the ESDF and is also known as the medial axis.
With some additional filtering and generation rules, the GVD can be used to create sparse paths through the environment which maximize obstacle clearance and preserve topological connections within the space.

However, these approaches have scarcely been extended to 3D for robot planning. 
Hoff \etal \cite{hoff2000interactive} and Foskey \etal \cite{foskey2001voronoi} explored using GVDs in 3D (generated by computing 2D slices of the GVD at various heights on GPU) given perfect CAD mesh data of the environment for path planning.
To the authors' best knowledge, no work has generated and used 3D GVDs for path planning on noisy, real-world data.

\begin{figure}[tb]
  \centering
  \includegraphics[width=1.0\columnwidth,trim=0 0 0 0 mm, clip=true]{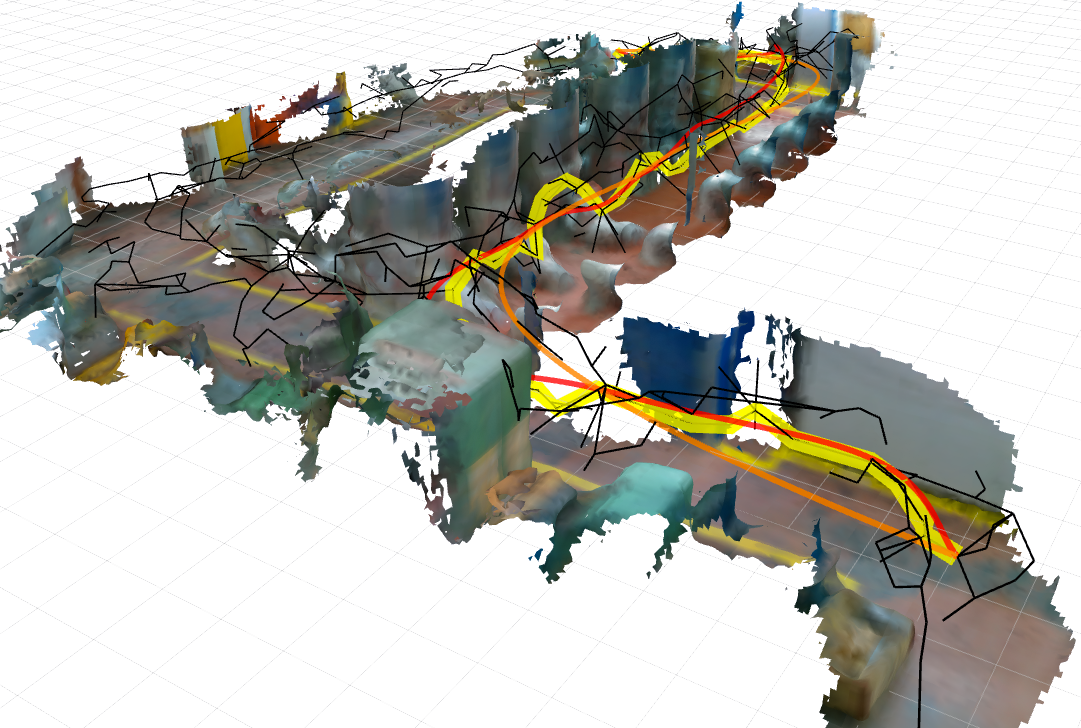}
  \caption{Sparse graph skeleton (black) and paths planned through a map created from on-board RGB-D sensing on an MAV in a machine hall. Yellow shows the initial path planned through the skeleton, orange is a dynamically-feasible path based on the initial path, and red is re-optimized to avoid collisions with the environment.}
  \label{fig:teaser}
\end{figure}

The aim of this paper is to address this research gap, and explore methods to generate descriptive, topology-preserving sparse graphs of 3D space that are suitable for global path planning for Micro-Aerial Vehicles (MAVs).
The main use-case we target is that of multi-session flying for industrial inspection or search and rescue applications.
In our scenario, the MAV has an initial exploration mission to build a map of the space in which it will operate.
This map is then refined and processed off-line, and a sparse topological graph representing the environment is created.
This graph can then be queried for very fast initial global plans to return to any point in the previously-explored map, and can be further refined with polynomial optimization techniques presented in our previous work~\cite{oleynikova2016continuous-time,oleynikova2017safe} and in \reffig{fig:teaser}.

Our method starts from a dense voxel map of 3D ESDF values, built using \textit{voxblox}~\cite{oleynikova2017voxblox}, extracts the 3D GVD or medial axis using methods inspired by graphics skeletonization literature~\cite{foskey2003efficient}, and creates a skeleton diagram by preserving only 1-voxel-thick edges of the medial axis.
We then fit a sparse graph, consisting of vertices and straight-line edges, to this underlying diagram, and ensure that it maintains connectivity.
Both the map construction and planning methods are made available open-source as part of \textit{voxblox}\footnote{\href{http://www.github.com/ethz-asl/voxblox}{\texttt{github.com/ethz-asl/voxblox}}}.

%We will show that this sparse graph is resistant to noise and resolution changes in the underlying map, and speeds up global planning by almost three orders of magnitude over RRT* and other methods.

%trim option's parameter order: left bottom right top

The contributions of this work are as follows:
\begin{easylist}[itemize]
  & a novel method of extracting thin skeleton diagrams from real, noisy sensor data in dense voxel grids (\refsec{sec:skeleton_diagram}),
  & a method of constructing sparse, straight-line connected graphs to use for rapid planning, which is resistant to noise and resolution changes (\refsec{sec:sparse_graph}),
  %& An evaluation of the effect noise and changing voxel sizes have on the structure of the sparse graph (\refsec{sec:results})
  & planning evaluations showing that our method can be orders of magnitude faster than other global planning methods (\refsec{sec:results}).
 %& Sparse graph and sample trajectory results on a map built in-flight from on-board MAV sensing
\end{easylist}

\section{Related Work}
In this section, we discuss existing robotics planning literature, with a focus on methods that exploit the topology of the scene.
We split our discussion into 2D methods and 3D methods.

\subsection{2D Planning}
%Planning in 2D topological graphs has been around for over 20 years.
Thrun was one of the first to show topological grid-based planning in 2D for ground robots~\cite{thrun1998learning}, where he built a topological map from Voronoi Graph, focusing on ``critical points" (doorways, etc.) to divide into topologically separate regions, and was able to show a three orders of magnitude speed-up over grid-based planning.

More recent work has focused on building signed distance fields and GVDs incrementally and in real-time.
Kalra \etal introduced the original dynamic brushfire algorithm, which is able to maintain the accuracy of the underlying signed distance field by using raise and lower wavefront propagation, and by keeping an updated list of GVD candidates~\cite{kalra2009incremental}.
Lau \etal further extended this method to create one-voxel-thin GVDs that better represented the topology of the underlying space~\cite{lau2013efficient}.

%Liu \etal extended the topological maps built from GVDs to also incorporate a concept of semantics; rather than attempting to faithfully represent the geometrical topology of the space, they focus on extracting room- and corridor-level understanding.

Fang \etal use 2D GVDs to inform sampling in 3D space for MAV navigation in indoor environments~\cite{fang20162d}.
They build a 2D GVD from a down-projection of the environment, and use this to overcome the difficulties that sampling-based methods have with narrow corridors and openings.
In contrast, our method actually builds a 3D GVD, fully capturing the geometry of the space.

\subsection{3D Planning}
We will cover three different categories of 3D planning: GVD-based 3D planning in CAD meshes, building topologically-aware maps based on something other than the GVD, and finally methods that explicitly model polygonal bounds of free-space regions for planning.

Hoff \etal presented a 2D but 3D-applicable method for building GVDs and then planning using potential fields in them. 3D is done with multiple 2D slices, and computed using graphics hardware~\cite{hoff2000interactive}.
The obstacles are from a labelled CAD model, and obstacles are defined per-object for the purposes of Voronoi boundaries (that is, a point belongs on the GVD if it is equidistant to two distinct \textit{objects}, rather than any surface boundaries).
Foskey \etal extended \cite{hoff2000interactive} to use a GVD for path planning directly in the graph, and if no solution through the GVD is found, use Probabilistic Roadmaps with sampling informed by the GVD.

Other methods for building topologically-aware graphs include SPARTAN, where a distance field is computed up to a certain distance from obstacles~\cite{cover2013sparse}, and connected in straight lines between distance field boundaries.
%Vertices in the graph are distance field ridges and the outside edges of distance computations around obstacles, and edges are only created at locations tangent to the distance field gradients at the vertices. \todo{rephrase}
This representation is well-suited for very sparse environments, as it allows shortcuts through large open-space regions, but does bug-algorithm-style planning around obstacles.
While their method is based on visibility graphs, we use Voronoi diagrams as our basis.

Another method is presented by Blochliger \etal, where SLAM-map landmarks projected into an occupancy map and then free-space is grown from the original robot map-inspection trajectory into mostly convex 3D clusters~\cite{blochliger2017topomap}. These clusters are then connected based on overlap to create a sparse navigation graph. The downside of this approach is that it heavily depends on the initial inspection trajectory rather than the inherent structure of the space.

Finally, another representation that has advantages for dynamics-aware planning is representing the environment as convex free-space regions.
A well-known method is IRIS~\cite{deits2015computing}, which iteratively breaks up a world into polygonal free-space regions from a seed point, but takes hours to compute on any non-trivial 3D environment.
Liu \etal attempt to overcome this speed limitation by building convex free space regions online, around an initial path found through A* or JPS graph search in a discretized space~\cite{liu2017planning}.
However, the final path will be a homotope of the initial A* path, so the free space regions can only be used for local refinement of the trajectory.
Similarly, Chen \etal exploit the structure of their obstacle map representation, Octomap, to grow and merge free-space axis-aligned cubes around an initial path and find solutions through this reduced graph, and then refine them using polynomial trajectory optimization~\cite{chen2016online}.

In contrast to all these methods, we compute the GVD-based topology of an entire noisy map, can be built on-line on-board an MAV, efficiently represents the underlying topology of a space, makes no assumptions about obstacle densities or using initial graph-search solutions.

\section{Skeleton Diagram Construction}
\label{sec:skeleton_diagram}
%\begin{easylist}[itemize]
%  & Generate ESDF
%  & Find GVD (some rules)
%  && $\theta$-SMA
%  && Maybe some pruning already at this stage
%  & Skeletonize GVD -> sparse skeleton graph
%  && Prune vertices
%  && Connect vertices with edges
%  && Find convex free space around edges, insert ``fake" vertices if necessary
%  && Re-construct missing edges from faces (maybe check how many neighbors belong to GVD)
%\end{easylist}
%  

\begin{figure}[tb]
  \centering
  \includegraphics[width=1.0\columnwidth,trim=0 0 0 0 mm, clip=true]{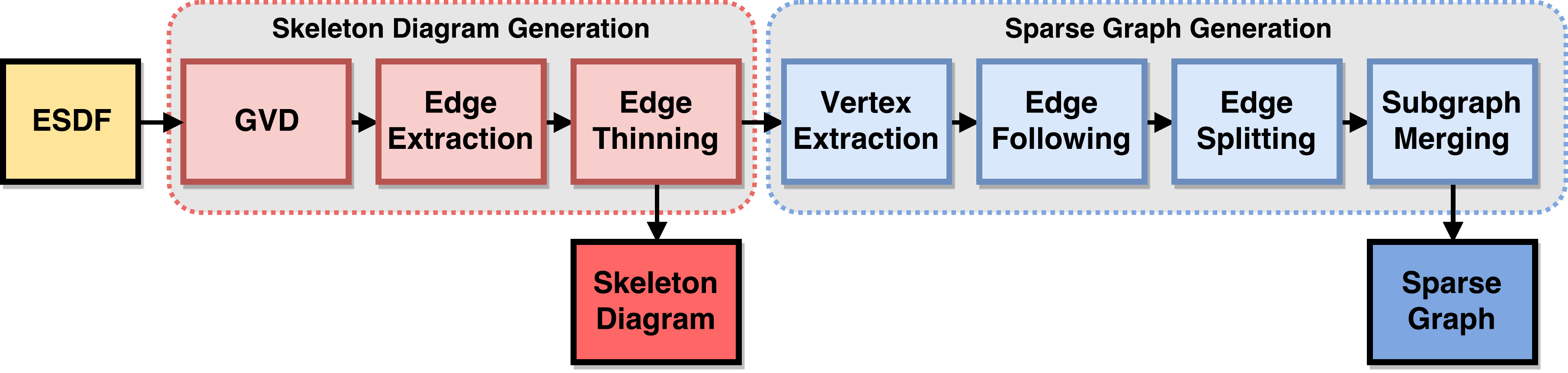}
  \caption{System diagram of the steps to generate the skeleton diagram (which is a dense voxel grid), and from there the sparse graph (which is an undirected graph with straight-line edges).}
  \label{fig:system_diagram}
\end{figure}
%trim option's parameter order: left bottom right top

\begin{figure}[tb]
  \centering
  \includegraphics[width=0.8\columnwidth,trim=0 0 0 0 mm, clip=true]{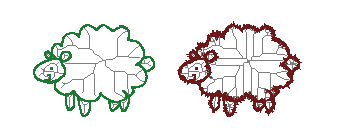}
  \caption{The effect of noise on the Voronoi diagram. The left image, uncorrupted by noise, has a very sparse diagram. The right image, generated with noisy sensor measurements, has many more edges in the diagram. Part of this work focuses on pruning the Voronoi diagram in the presence of heavy sensor noise in the map.}
  \label{fig:noise_voronoi}
\end{figure}
%trim option's parameter order: left bottom right top

Our method consists of two main components: first, building a one-voxel-thick \textit{skeleton diagram} in the voxel space, which preserves the topology and connectivity of the original space while having as few elements as possible, as can be seen in 2D in \reffig{fig:noise_voronoi}.
This is the closest analogue to the 2D topological maps used by Lau \etal~\cite{lau2013efficient}.
Second, generating a \textit{sparse graph} out of the diagram, which is no longer bound to the voxel space, and instead consists of graph vertices connected by straight-line edges.

The overall system diagram is shown in \reffig{fig:system_diagram}.
The general order of steps is to generate the Generalized Voronoi Diagram (GVD), or medial axis, of the space, extract its edges and vertices, thin the diagram, then create a sparse graph by following edges in this diagram, and reconnect disconnected subgraphs in the final graph.

\subsection{GVD Generation}
In order to generate the original Generalized Voronoi Diagram, we follow the common approach of finding ``ridges" in the signed distance function~\cite{tagliasacchi20163d, foskey2001voronoi}.
Finding the GVD is analogous to finding a discretization of the medial axis of the free-space, which has been well-studied in 2D  robotic planning problems~\cite{thrun1998learning, lau2013efficient}.
In 2D, the medial axis consists of connected lines that maximize the distance from obstacles, as shown in \reffig{fig:noise_voronoi}.
However, since the medial axis has one dimension less than the original shape, when this is generalized to 3D, the medial axis consists of (potentially curved) planes that maximize this distance.
To make the problem as sparse as possible for planning, we discard the planar representation and aim instead to find the medial curve skeleton, which consists of 1-voxel-thick lines in 3D. This is analogous to the edges and vertices of the 3D GVD, discarding the faces.

We begin with the same general approach as Foskey \etal~\cite{foskey2001voronoi} to find all the points that belong on the medial axis.
To do this, we iterate over all voxels in the ESDF that are in free space, and compare their parent direction with the parent direction of their 6-connectivity neighbors.
Each voxel stores the direction to its ``parent" point on the object surface (the closest occupied point to the voxel).
Note that we could also follow the 2D approach of Lau \etal~\cite{lau2013efficient} and store the GVD candidates from the termination of the wavefront to avoid having to check all the voxels in the map.

When the medial axis is constructed from CAD-generated data, where it is known which surface belongs to which object, it is sufficient to check if the neighbors have different parents or basis objects~\cite{hoff2000interactive}.
However, with noisy real-world data and discretized map representations, as show in \reffig{fig:noise_voronoi}, it is difficult to tell which points belong to which objects, and noise on the surface boundary creates many spurious medial lines.

To combat this problem, we adapt the $\theta$-SMA approach~\cite{foskey2003efficient}, which creates a simplified medial axis (SMA) based on a minimum $\theta$ angle between basis points.
In our discretized ESDF, we compute whether a point belongs on the medial axis using:

\begin{equation}
\frac{\vec{n}_p + \vec{n}_d}{\norm{\vec{n}_p + \vec{n}_d} } \bullet \frac{\vec{v}_p}{\norm{\vec{v}_p} } < \cos{ \theta },
\end{equation}
where $\vec{v}_p$ is the parent direction of the current voxel, $\vec{n}_d$ is the direction from the neighbor voxel to the current voxel, and $\vec{n}_p$ is the parent direction of the neighbor voxel.

Passing this check means that the point has at least two \textit{distinct} basis points on object boundaries.

%\begin{eqnarray}
%dp &= \norm{\textrm{neighbor voxel parent direction} \\ 
%  & + \textrm{direction to neighbor}} \\
%&\dot \norm{\textrm{voxel parent dir}}
%%dp = \norm{\textrm{neighbor voxel parent direction} + \textrm{direction to neighbor}} \dot \norm{\textrm{voxel parent dir}}
%\end{eqnarray}
%
%

%Start out with approach of Foskey \etal~\cite{foskey2001voronoi}, though we have additional problems: discretization error. Generate with 6-connectivity. 

\subsection{Edge Extraction}
\begin{figure}[tb]
  \centering
  \begin{subfigure}[b]{0.48\columnwidth}
    \centering
    \includegraphics[width=1.0\columnwidth,trim=0 0 0 0 mm, clip=true]{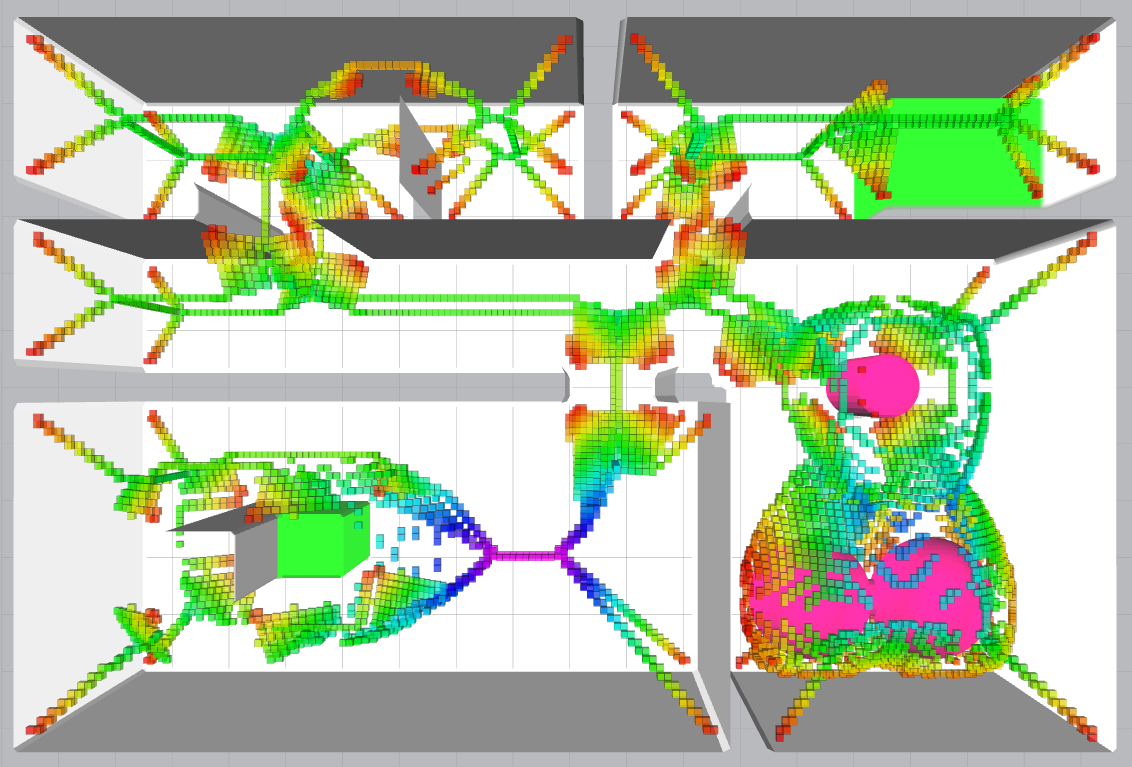}
    \caption{}
    \label{fig:edges_medial}
  \end{subfigure}
  \begin{subfigure}[b]{0.48\columnwidth}
    \centering
    \includegraphics[width=1.0\columnwidth,trim=0 0 0 0 mm, clip=true]{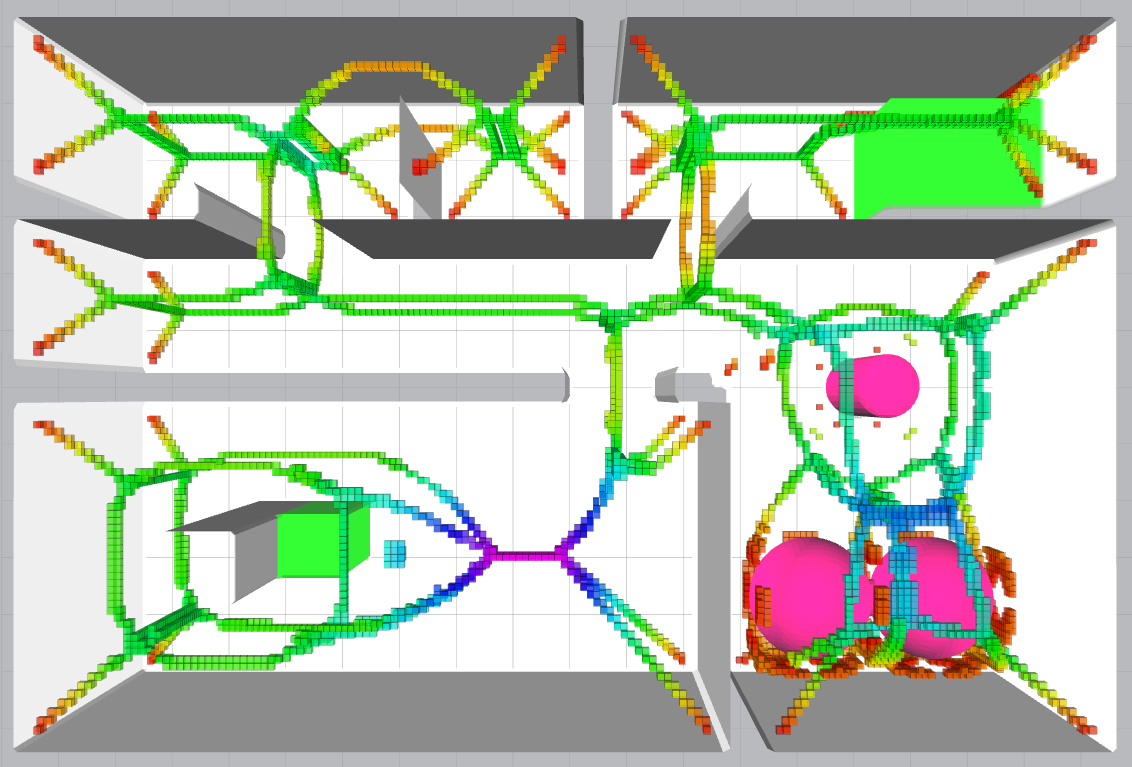}
    \caption{}
    \label{fig:edges_neighbor}
  \end{subfigure}
  \begin{subfigure}[b]{0.48\columnwidth}
    \centering
    \includegraphics[width=1.0\columnwidth,trim=0 0 0 0 mm, clip=true]{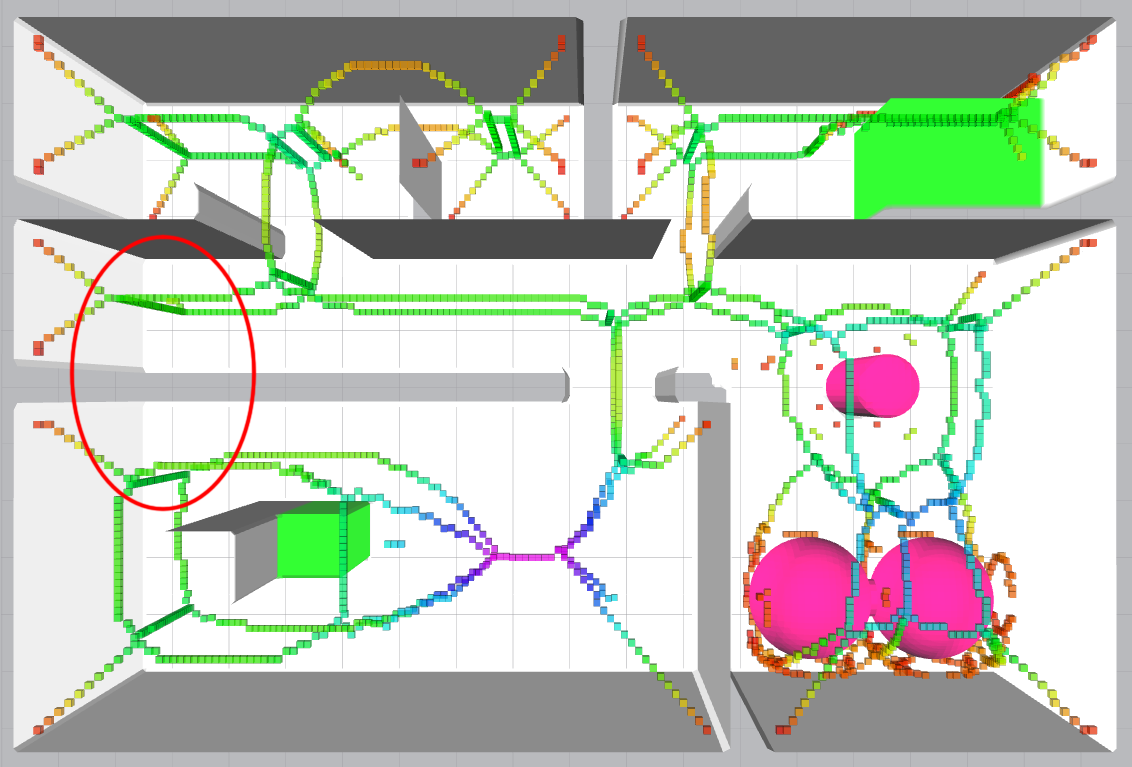}
    \caption{}
    \label{fig:edges_thin_noend}
  \end{subfigure}
  \begin{subfigure}[b]{0.48\columnwidth}
    \centering
    \includegraphics[width=1.0\columnwidth,trim=0 0 0 0 mm, clip=true]{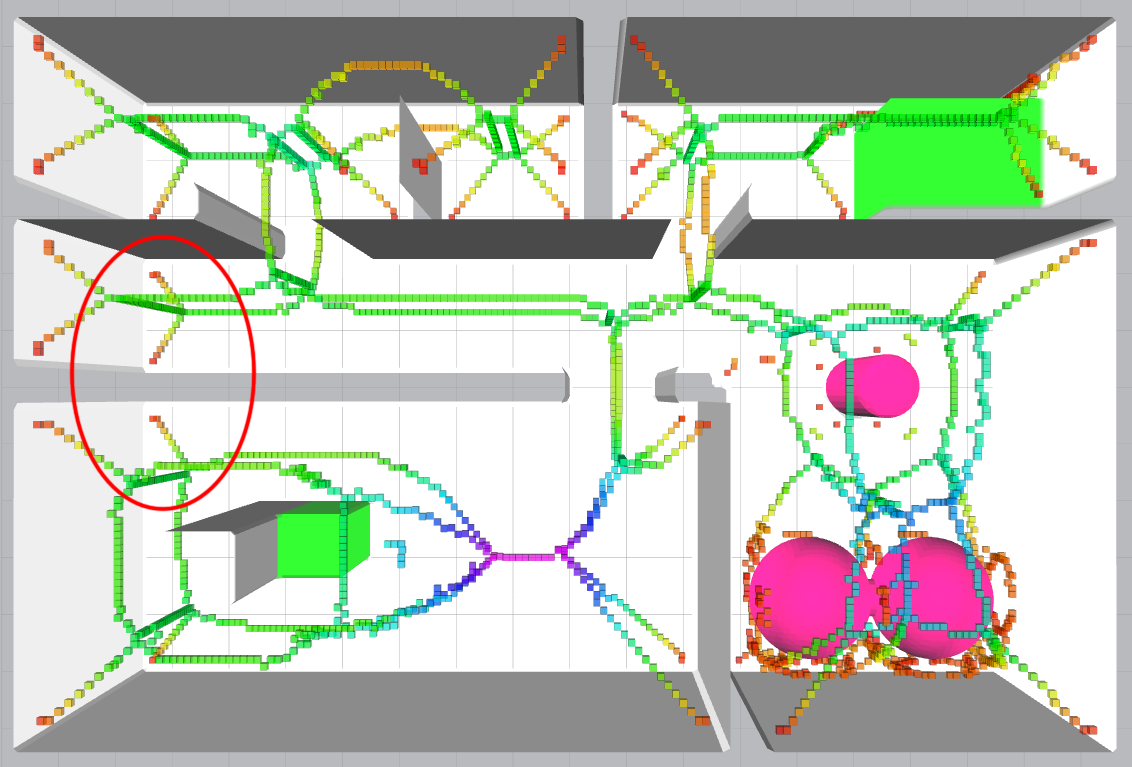}
    \caption{}
    \label{fig:edges_thin_end}
  \end{subfigure}
  \caption{Different methods of generating the edges of the medial axis/GVD. \subref{fig:edges_medial} shows the edges generated by counting the number of basis points, \subref{fig:edges_neighbor} shows edges generated by number of neighbors that are also on the medial axis, \subref{fig:edges_thin_noend} shows edges from \subref{fig:edges_neighbor} thinned using topology-preserving erosion techniques, and finally \subref{fig:edges_thin_end} shows the same technique applied but without extended end-point checks, note especially the edges preserved in the red circle. Color represents the distance to the obstacle.}
  \label{fig:edge_definitions}
\end{figure}
%trim option's parameter order: left bottom right top

Now that we have all the points that belong on the medial axis, we need to classify them as edges, vertices, and faces, as we aim to only keep the one-voxel-thin lines that sparsely describe the topology of the space.

In a perfectly modeled environment, the edges of a 3D GVD are defined as having three basis points, and vertices as having four~\cite{hoff2000interactive}. 
However, due to discretization error, and the actual Voronoi boundaries falling between voxels, some edges disappear and spurious edges appear by this definition, as shown in \reffig{fig:edges_medial}.

Since the Voronoi boundaries are also edges of the faces of the medial surface, we instead choose a method more stable to discretization error and noise in the boundaries: extracting edges of the medial axis, to create medial lines.
We use the metric of how many of the voxel's 26-connected neighbors belong to the medial axis to determine whether it is an edge.
We preserve all voxels with at least 18 neighbors, which creates the thick but complete skeleton in  \reffig{fig:edges_neighbor}.

However, even this definition is not free of errors introduced by discretization, especially depending on the size of the voxels.
Lau \etal suffered the same problem in their 2D GVD, where many lines are two-voxels-thick, and create spurious connections.
Their solution was to introduce two connectivity templates, which all pixels on the GVD were verified against to ensure that they were necessary to maintain the topology of the graph.
However, since 3D topology is much more complex, it is not sufficient to verify our edge voxels against connectivity templates; we must instead do a topology- and connectivity-preserving 3D thinning operation, described in the section below.

%Goals:
%\begin{easylist}[itemize]
%  & Restore connectivity between vertices using edges
%  & Similar structure despite different voxel resolution (overcome discretization error)
%  & Have a figure about pre-repair and post-repair
%  & Maybe have how to plan in this space already? (A* search...)
%  & Use 26-connectivity patterns to generate edges
%  && Then get vertices out of this
%  & Compare to doing this by Foskey approach~\cite{foskey2001voronoi}, which is very sensitive to voxel size (example figure?)
%\end{easylist}

\subsection{Thinning}
\begin{figure}[tb]
    \centering
  \begin{subfigure}[b]{0.48\columnwidth}
    \centering
    \includegraphics[width=1.0\columnwidth,trim=0 0 0 0 mm, clip=true]{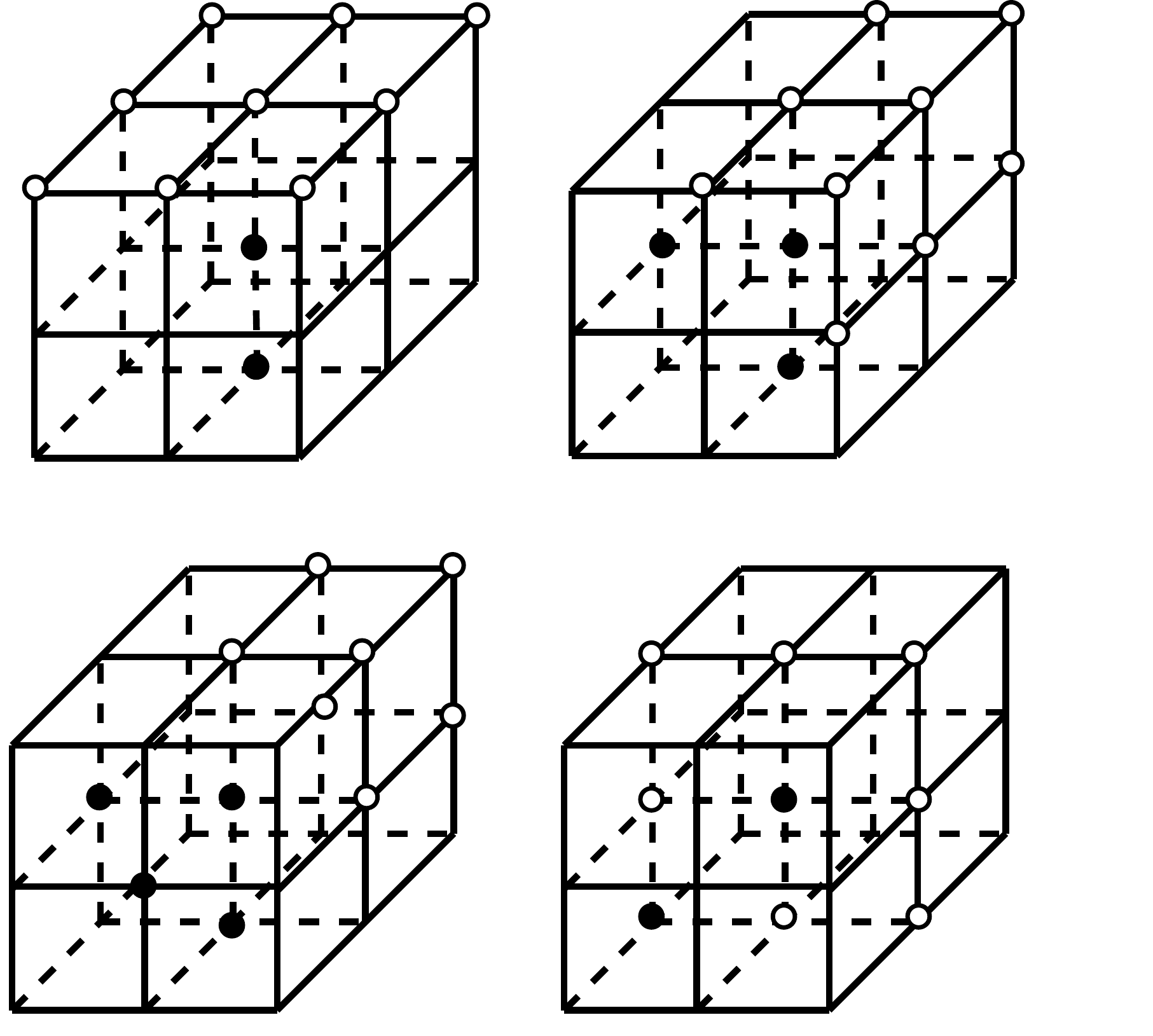}
    \caption{}
    \label{fig:deletion_templates}
  \end{subfigure}
  \begin{subfigure}[b]{0.43\columnwidth}
    \centering
    \includegraphics[width=1.0\columnwidth,trim=0 0 0 0 mm, clip=true]{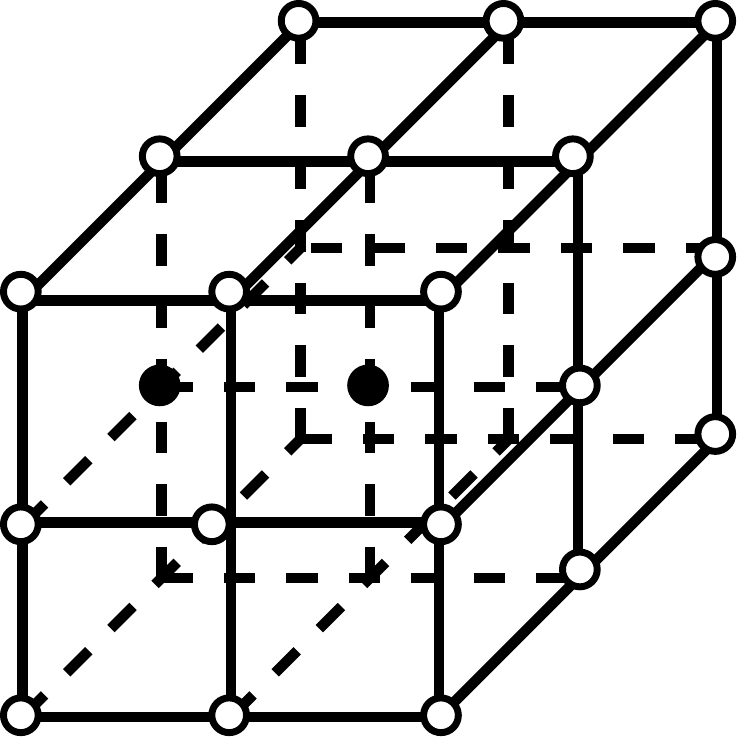}
    \caption{}
    \label{fig:corner_template}
  \end{subfigure}
  \caption{Voxel templates for thinning, \subref{fig:deletion_templates} is the 4 deletion templates from \cite{she2009improved}, and \subref{fig:corner_template} is our template to check whether a point is a 6-connected corner or not. Black points indicate foreground points (in our case, GVD edge points), white is background (non-edge points), and unlabelled match both foreground and background points.}
  \label{fig:thinning_templates}
\end{figure}
%trim option's parameter order: left bottom right top

In order to remove spurious double lines, we refer to digital topology literature which builds 3D skeletons out of discretized shapes using recursive thinning~\cite{she2009improved, ma1996fully, lee1994building}.

We follow the general approach described in Lee \etal, which would also allow parallelization of the thinning operation~\cite{lee1994building}.
First, all voxels that are part of the shape are checked against one of several deletion templates.
We choose to use the patterns from She \etal~\cite{she2009improved}, due to their simplicity and not requiring more than the $3\times3$ neighborhood for every voxel, shown in \reffig{fig:deletion_templates}.
The neighborhood of the center voxel is evaluated against these templates.
In the templates, a point may be either foreground (black, edge or vertex neighbor), background (white, face or non-GVD neighbor), or ``don't care" (unmarked).
``Don't care" points can match against either value.

Once a point is shown to match one of the templates (or any of its $90\deg$ rotations or mirror operations), it can be removed if it is a \textit{simple} point: that is, if the connectivity of its neighbors would be preserved without it~\cite{lee1994building}, and not an end-point: has more than one 26-connected neighbor.
This definition is both for groups of connected foreground voxels (connected using 26-connectivity) and connected background voxels (connected using 6-connectivity).
Therefore, to verify that a point is simple, we need to verify that the number of connected components in its neighborhood remains the same without it.
We do this using the octree adjacency tests proposed in Lee \etal~\cite{lee1994building}.

However, since our edge-extraction heuristic produces 6-connected edges, removing all simple points that are not end-points leads to incorrect removal of some diagonal edges.
This is due to the simplicity of the end-point test: a 6-connected end-point actually has \textit{2} 26-connected edges.
Classifying any point that only has one 6-connected neighbor leads to incorrect preservation of many other points.
(This is not a problem encountered in medial skeleton construction through thinning, since the thinning is applied recursively and should always lead to only 26-connected components.)

To combat this issue, we extend the end-point test with a ``corner template", shown in \reffig{fig:corner_template}.
A point is considered an end-point if it has only one 26-connected neighbor, or if it has no more than one 6-connected neighbor and matches the template (and its rotations and mirrors) in \reffig{fig:corner_template}.

The results of this thinning operation is a fully-connected, one-voxel-thin skeleton, shown in \reffig{fig:edges_thin_end}.
This state of the map is referred to as the \textit{skeleton diagram} for the remainder of the paper, and is analogous to the GVDs used for planning in 2D literature.

%\cite{she2009improved} [She 2009]: Improved/simplified templates to use for 3D thinning. Sufficient conditions for deletion (simple point and doesn't introduce a hole), though no good algorithmic way to verify them.

%\cite{lee1994building} [Lee 1994]: Great reference for thinning for medial skeleton extraction. Uses the 1-thick keleton to make a sparse graph to use for casting/forging parts analysis. Has a great description of the overall algorithm (not template based) and how to implement simple point checking.

\section{Sparse Graph Generation}
To further sparsify the problem, we approximate the skeleton diagram with a sparse graph, using the steps described in this section.

\label{sec:sparse_graph}
\subsection{Vertex Extraction}
\begin{figure}[tb]
  \centering
  \begin{subfigure}[b]{0.48\columnwidth}
    \centering
    \includegraphics[width=1.0\columnwidth,trim=0 400 400 0 mm, clip=true]{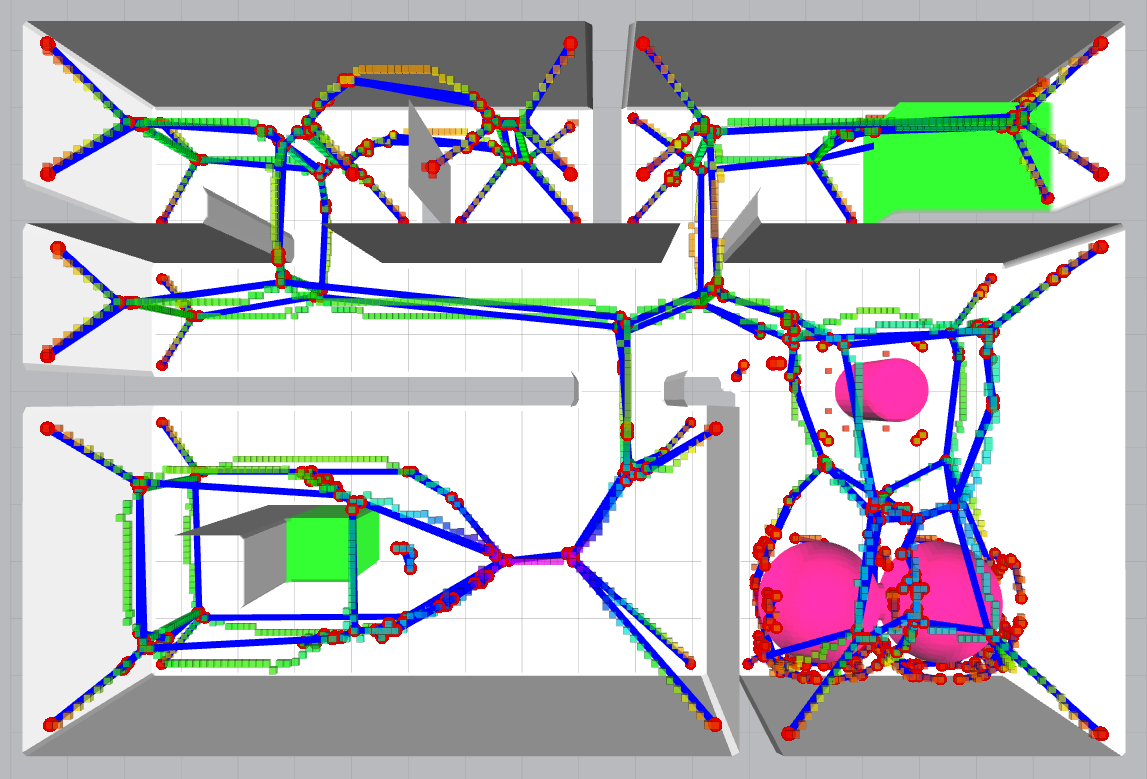}
    \caption{}
    \label{fig:vertices_no_prune}
  \end{subfigure}
  \begin{subfigure}[b]{0.48\columnwidth}
    \centering
    \includegraphics[width=1.0\columnwidth,trim=0 400 400 0 mm, clip=true]{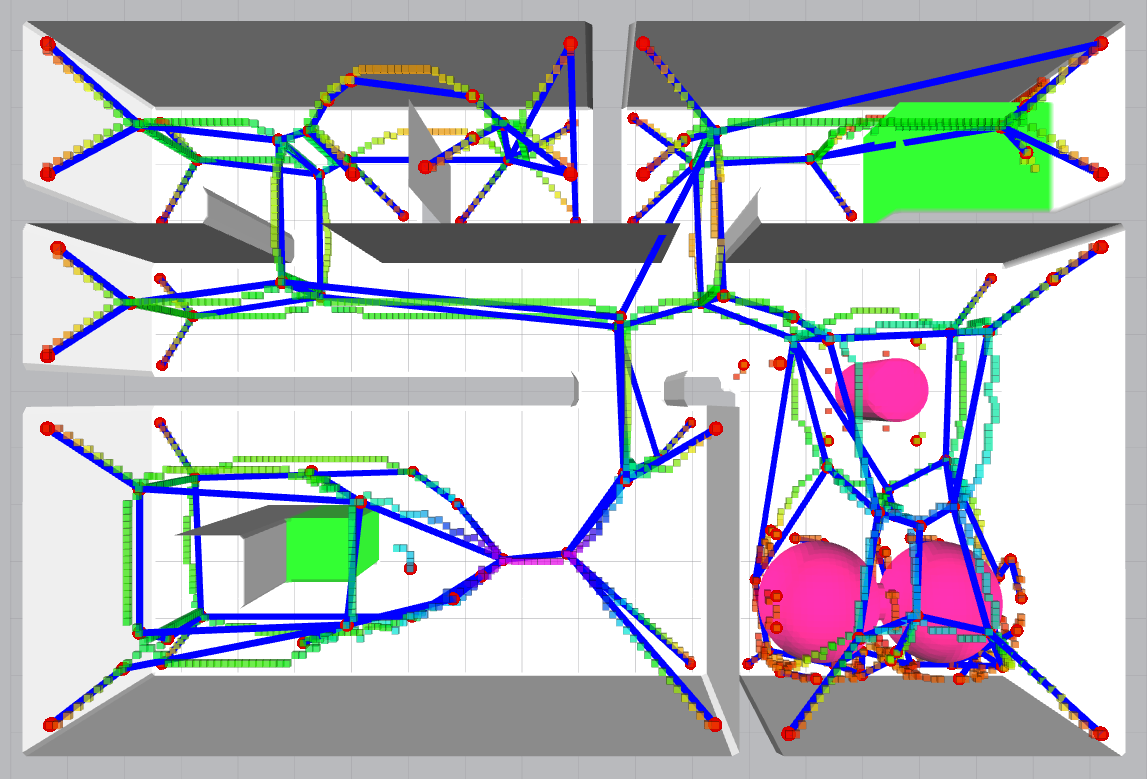}
    \caption{}
    \label{fig:vertices_prune}
  \end{subfigure}
  \begin{subfigure}[b]{0.48\columnwidth}
    \centering
    \includegraphics[width=1.0\columnwidth,trim=0 400 400 0 mm, clip=true]{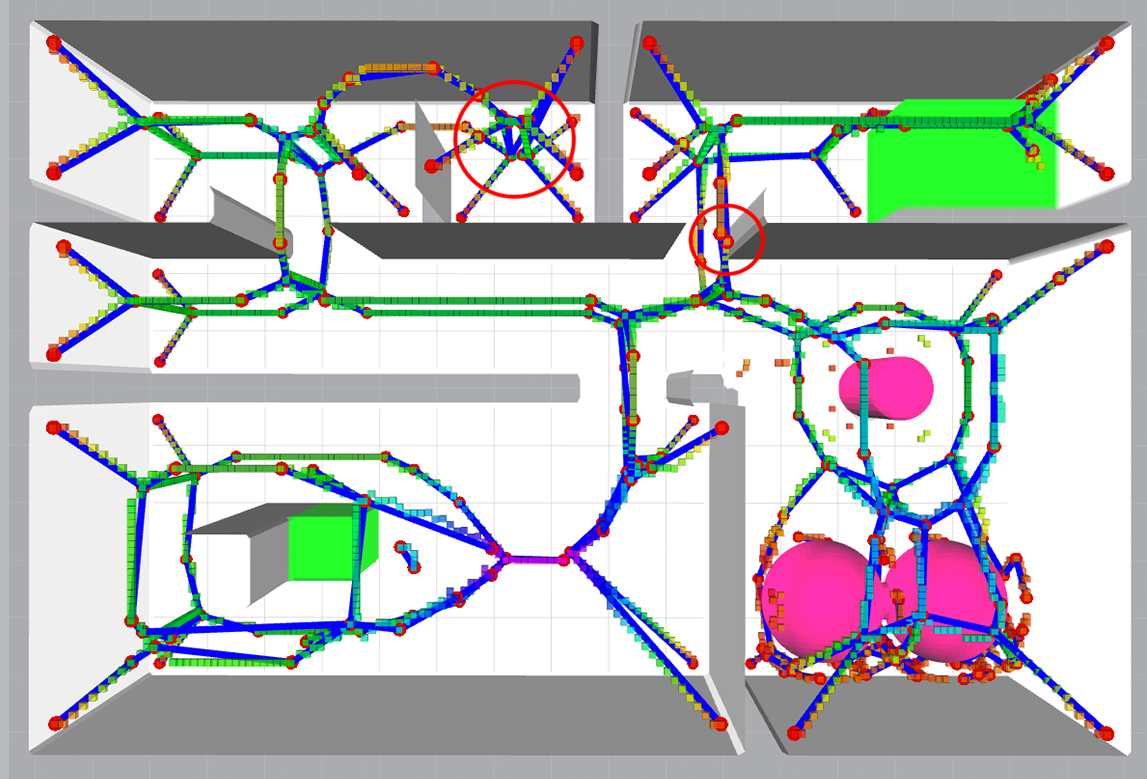}
    \caption{}
    \label{fig:vertices_split}
  \end{subfigure}
  \begin{subfigure}[b]{0.48\columnwidth}
    \centering
    \includegraphics[width=1.0\columnwidth,trim=0 400 400 0 mm, clip=true]{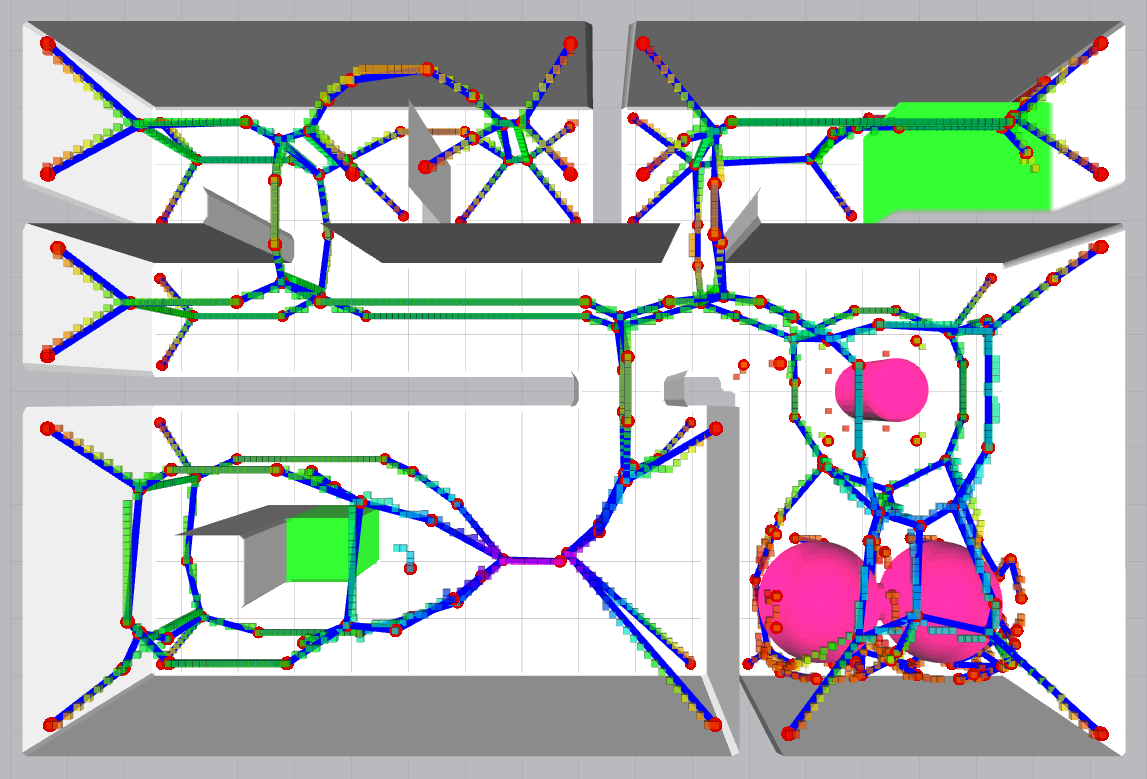}
    \caption{}
    \label{fig:vertices_split_and_match}
  \end{subfigure}
  \caption{Steps in the process of vertex generation, with red circles as sparse graph vertices and blue connecting lines as sparse graph edges, overlaid on top of the GVD edges. \subref{fig:vertices_no_prune} shows a connected sparse graph, without any pruning of vertices. \subref{fig:vertices_prune} shows the result with k-D tree pruning; as can be seen, some edges pass through obstacles. \subref{fig:vertices_split} shows the results of splitting the edges any time they deviate too far from the straight line by inserting a new vertex, and \subref{fig:vertices_split_and_match} shows the results if the splitting also attempts to match to a nearby vertex (removing duplicated near-by vertices seen in \subref{fig:vertices_split}).}
  \label{fig:vertices}
\end{figure}
%trim option's parameter order: left bottom right top

After our skeleton is only one-voxel-thick, we are able to extract vertices very simply: finding all edge voxels that have 1 or more than 3 26-connected neighbors that are edges.
The results of this are shown in \reffig{fig:vertices_no_prune}, with red circles indicating vertices.
As can be seen, there are many redundant nearby vertices due to this definition.
While they are technically correct, they clutter the sparse graph and add no new topologically-distinct paths.

In order to ensure that the graph we build in the next stage is as sparse as possible, we prune the vertices in the GVD.
We build a k-D tree of the nearest vertices, and for each vertex in the GVD, find all other vertices that are within a pruning radius $r_\textrm{prune}$.
From these vertices, the one with the largest distance to obstacles is retained, and all others are removed.
The results of this operation are shown in \reffig{fig:vertices_prune}.

\subsection{Edge Following}
The next step is to trace straight-line edges between the sparsified vertices obtained from the previous section.
The goal is to create a graph that is independent of the resolution of the underlying map, and is significantly faster to search through as all connectivity information is pre-computed.

We start with the filtered diagram vertices from the previous step.
Each vertex is assigned a vertex ID, which is also marked in the diagram, and inserted into a map of vertex IDs to vertex information.
For each vertex, we attempt to follow the diagram edges to find connections to other vertices.

This is done as follows: first, for each vertex, we check its 26-connectivity neighbors for edges.
For each edge, we recursively follow it through the diagram, by checking all its neighbors and preferring those most in line with the current direction of the edge:
\begin{equation}
\min{\big((\vec{v}_d - \vec{r}_d) \bullet -\vec{n}_d}\big),
\end{equation}
where $\vec{v}_d$ is the direction taken to get to the current voxel, $\vec{r}_d$ is the direction from the vertex, and $\vec{n}_d$ is the direction from the current voxel to the checked neighbor.

Directions that are not followed are stored in a stack (FILO), and in case there are no more viable adjacent neighbors, the latest candidate is popped off the stack.

This procedure is followed until a vertex voxel is reached, then its corresponding entry in the sparse graph is found, and an edge connecting the originating voxel and the new one is created.

The results of the complete graph generation procedure, including edges shown as straight blue lines, is shown in \reffig{fig:vertices_prune}.
As can be seen, some edges do not sufficiently represent the structure of the underlying diagram: for instance, some curved edges are represented as straight lines and as a result pass directly through an obstacle.
This is obviously undesirable in a graph used for planning.

%\begin{easylist}[itemize]
%  & Filter vertices, maybe based on k-D tree and min distance between them, pick one with biggest distance.
%  && Non-max supression in local area?
%  & From filtered vertices, follow edges to the nearest vertex.
%  & If graph isn't fully connected (or anyway?), try to search for connections in edge graph using Djikstra or something.
%  & Add vertices when the edges tighten/increase or change direction sharpy.
%  & In fact maybe compare edge-following vs. k-D tree and Djikstra search...
%\end{easylist}

\subsection{Edge Splitting}
In order to prevent edges going through non-free space, we split edges that deviate too far from the straight-line path.
This is done as follows: first, for every edge in the graph, we use diagram A* (described below in \refsec{sec:diagram_astar}) to find the shortest path in the diagram from the start vertex to the end vertex.
Every point in this diagram edge is then projected onto the straight-line between the start and the end vertex, and its distance to the line is calculated.
If the maximum distance along this edge exceeds a threshold (for instance, we used twice the voxel size), then a new candidate vertex is created at the point on the diagram with maximum distance.

Making every candidate vertex into a real vertex gives the results in \reffig{fig:vertices_split}: while the edges now match the shape of the diagram edges, there are some cases where unnecessary vertices are created: that is, locations where there is already a vertex nearby that the edge should have been connected to instead.
To counteract this case, we add another check before adding a candidate vertex as a new vertex: if there is another vertex within $r_\textrm{prune}$ of the candidate, then use diagram A* to verify that a valid connection exists to both the start and the end vertices.
This vertex candidate is considered \textit{only} if connecting to it reduces the maximum distance to the straight-line.

The results of this final step are shown in \reffig{fig:vertices_split_and_match}, where it can be seen that significantly fewer extra vertices are made, and shorter edges to existing vertices are established.

\subsection{Disconnected Sub-Graph Repair}
In some cases for large, cluttered maps, such as the maze we will use in \refsec{sec:planning_results}, some edges that are present in the diagram are not re-connected correctly in the sparse graph due to discretization error.
This leads to multiple disconnected subgraphs.

We solve this problem by first determining the number of subgraphs and which vertices belong to them.
For each vertex in the graph, if it is unlabelled with a subgraph ID, we assign it a new subgraph ID and perform the flood fill algorithm to label all the vertices that belong to its subgraph.
The flood fill algorithm is a simple recursive depth first search, where we set the label of each vertex, follow all the edges from that vertex, and continue until no un-labelled vertices that are reachable by edges remain.

After each vertex is labelled, we perform two steps: first, remove all subgraphs that contain only one vertex, as these are not helpful in planning.
Second, attempt to connect all the remaining subgraphs to each other.

We do this by using A* through the skeleton diagram.
We select the first labelled vertex in two disconnected sub-graphs, and attempt to find a path between them along the underlying skeleton diagram.
If A* is able to find a path, we trace back through it, checking every voxel along this path.
For every voxel that is assigned a sparse-graph vertex ID, we record its ID and label.
When the labels between two consecutive vertices change, we add an edge between those vertices and perform flood fill to re-label the new connected graph.

Note that this approach may connect more than just the start and end subgraph, if other disconnected subgraphs are encountered on the path.
All new added edges are again checked for straight-line connectivity with the method described in the previous section.

We refer to the final state of this graph as the \textit{sparse graph} for the purposes of planning.

\section{Planning Algorithms}
\label{sec:planners}
In this section, we will describe various ways to use the information contained in the ESDF, skeleton diagram, and sparse graph for path planning.
These methods will be evaluated and compared in the results in \refsec{sec:planning_results}.
For each of these methods, we model the MAV as a sphere with a fixed radius. The GVD was generated with a minimum distance of this radius, so only valid traversible nodes are present in the graph.

For the first four planning methods, we focus on the goal of quickly finding a feasible initial path through the space, considering non-collision with obstacles as the only requirement.
This initial path will then be refined to be a smooth, dynamically-feasible trajectory for the MAV using polynomial splines and the property of differential flatness, as described in \refsec{sec:trajectory} below.

\subsection{A* through ESDF}
The simplest method to find a path through the space is to run A*~\cite{hart1968formal} on the ESDF voxels.
For each voxel, starting from the start location, we expand its 26-connected neighbors and consider them part of the graph if their ESDF distance is greater than the robot radius.
We use the straight-line distance to goal as an admissible heuristic, and accumulate voxel adjacency distances.

Though this approach is resolution-complete, it becomes prohibitively expensive in larger spaces or smaller voxel sizes due to the massive size of the graph.

\subsection{A* through Skeleton Diagram}
\label{sec:diagram_astar}
A much faster approach is to search only through the skeleton diagram.
%Since we have shown that the skeleton diagram preserves the topology and connectivity of the space, any feasible paths should be also represented in the diagram.
%This approach will not find minimum-lengths paths, as it maximizes clearance.
The heuristics and costs remain the same as in the ESDF A*, but a voxel is considered a valid neighbor if it is an edge or a vertex in the skeleton diagram.

One important consideration is that the start and end points may not necessarily be on the diagram.
To counter this, we start an A* search through the ESDF from the start toward the goal, and abort as soon as it expands a vertex that is also on the diagram.
We also start the same search backwards: from the goal toward the start.
These two searches quickly find the start and end points on the diagram.
%The final path consists of a concatenation of the path to the diagram start, the path along the diagram, and the path from the diagram to the goal.

\subsection{A* through Sparse Graph}
The final speed-up is to traverse the sparse graph rather than the underlying diagram, as the graph keeps a mostly consistent size regardless of voxel size or noise level, as shown in \refsec{sec:map_results}.

We find the sparse graph vertices closest to the start and end points by searching for the nearest neighbors in a pre-built k-D tree structure.
We then use A* to traverse the edges toward the goal vertex.
%, where the heuristic is again straight-line distance to the goal and the cost is the accumulation of straight-line distances along the edges.

\subsection{RRT through ESDF}
We also compare to a more traditional method of global planning, based on sampling-based Rapidly-Exploring Random Trees (RRTs).
We evaluate both RRT*~\cite{karaman2011sampling}, which is probabilistically-optimal (that is, it is guaranteed to find the optimal solution given infinite execution time) and RRT Connect~\cite{kuffner2000rrt}, which has no optimality guarantees but in practice quickly finds a sub-optimal feasible path.
In both cases, we treat unknown space as occupied and do collision-checking directly in the ESDF.
%A point is considered valid if its ESDF distance is greater than the robot radius.

Unlike the search-based methods described above, which exhaustively search a graph toward the solution, RRT-based methods sample feasible points in the planning space and attempt to connect them to the existing tree.
Once the goal state has been sampled and connected to the tree, the tree is traversed to get the final path.
In RRT Connect, the tree is grown bi-directionally from both the start and goal.
In RRT*, the initial path continues to be refined with shorter path-length solutions until a time limit is reached.

\subsection{Dynamically-Feasible Trajectory Generation}
\label{sec:trajectory}
Finally, given an initial straight-line path from any of these methods, we need to create a trajectory that the MAV can dynamically follow.
We exploit the property of differential flatness to plan dynamically feasible polynomial splines.
Since the splines may deviate from the initial straight-line solutions and no longer be collision-free, we use local trajectory optimization using the ESDF distances as collision costs to iteratively ``push" the final trajectory out of collision, as described in our previous work~\cite{oleynikova2016continuous-time,oleynikova2017safe}.

%\section{System and Planning}
%\begin{easylist}[itemize]
%  & Localization map generation
%  && Run rovioli, save resources
%  && Run bundle-adjustment
%  && Re-generate localization map and TSDF together
%  & Skeleton generation
%  && Generate skeleton using method above
%  & Global planning
%  && Use A* search to navigate through regions in the graph
%  && Within convex free space regions, use the convex space bounds to do trajectory generation
%  & Local planning
%  && Load skeleton or previous map as prior
%  && Replan when in collision with new map
%\end{easylist}

\section{Experiments}
\label{sec:results}
In this section aims to validate our method on simulated and real-world data and analyze the effect of noise and voxel size on the resulting sparse graph.
We also compare the different presented planning approaches, and finally show results on real maps from an MAV flight with generation of dynamically-feasible paths.

\subsection{Sparse Graph Construction}
\label{sec:map_results}

\begin{figure}[tb]
  \centering
  \settoheight{\tempdima}{\includegraphics[width=.30\columnwidth]{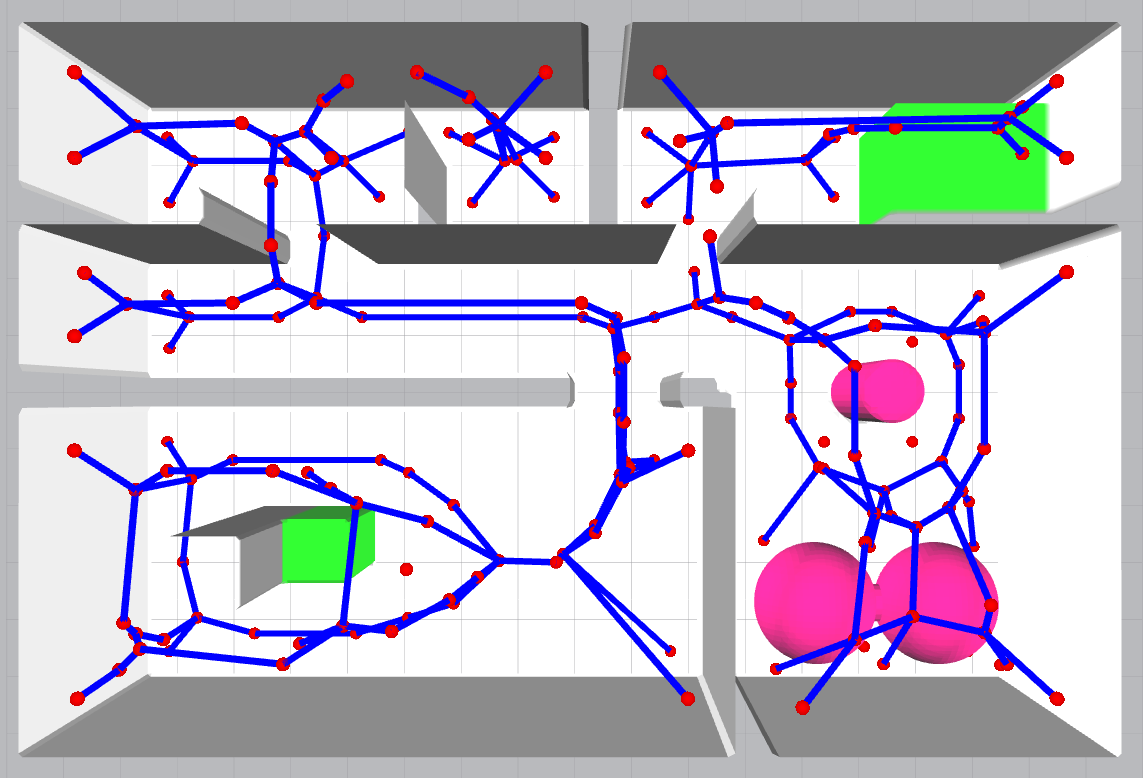}}%
  \centering\begin{tabular}{@{}c@{ }c@{ }c@{ }c@{}}
    % &0.02 m & 0.10 m & 0.20 m \\[-1.5ex]
    \rowname{Sim. Map}&
    \includegraphics[width=.30\columnwidth,trim=000 000 0 0 mm, clip=true]{figures/graph_results/10_sim.png}&
    \includegraphics[width=.30\columnwidth,trim=000 000 0 0 mm, clip=true]{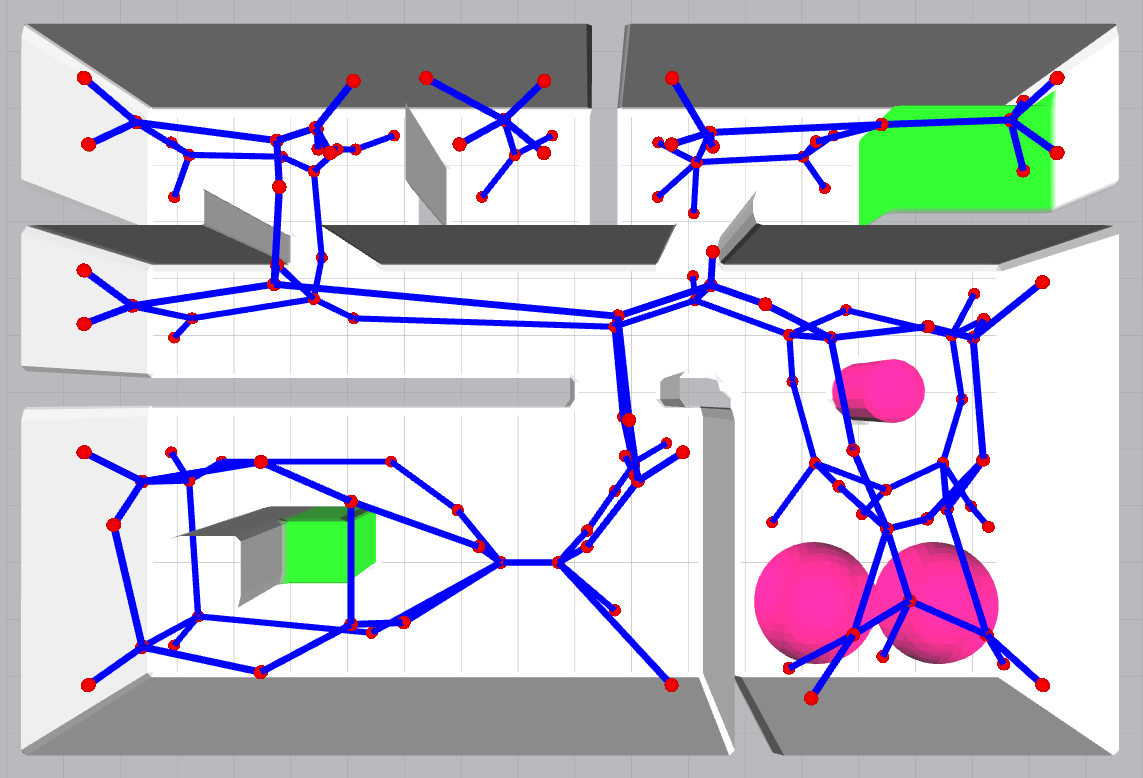}&
    \includegraphics[width=.30\columnwidth,trim=000 000 0 0 mm, clip=true]{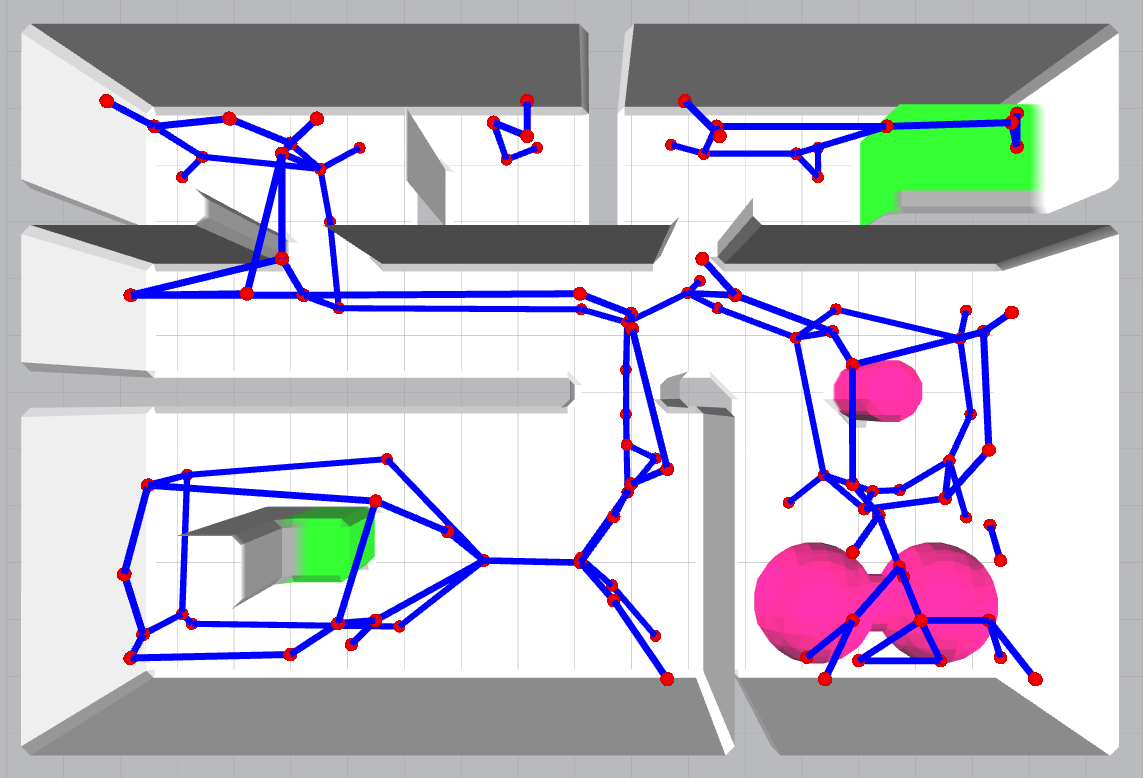}\\
    \rowname{$\sigma = 0$}&
    \includegraphics[width=.30\columnwidth,trim=000 000 0 0 mm, clip=true]{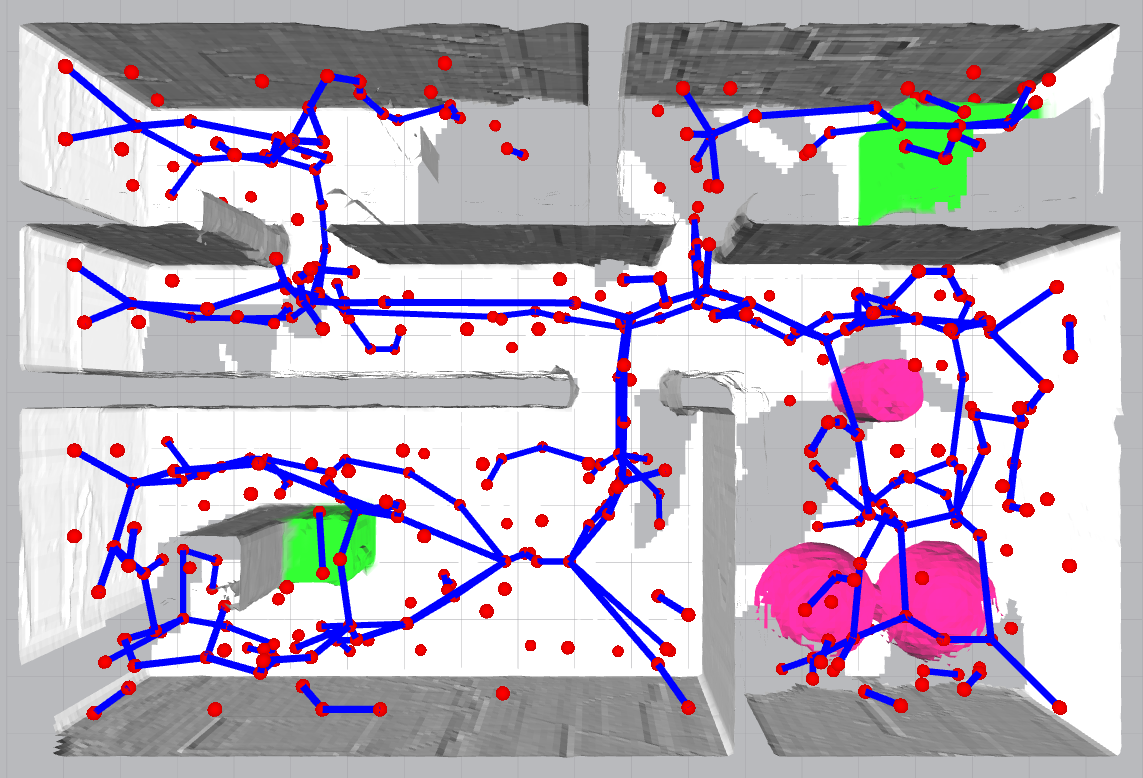}&
    \includegraphics[width=.30\columnwidth,trim=000 000 0 0 mm, clip=true]{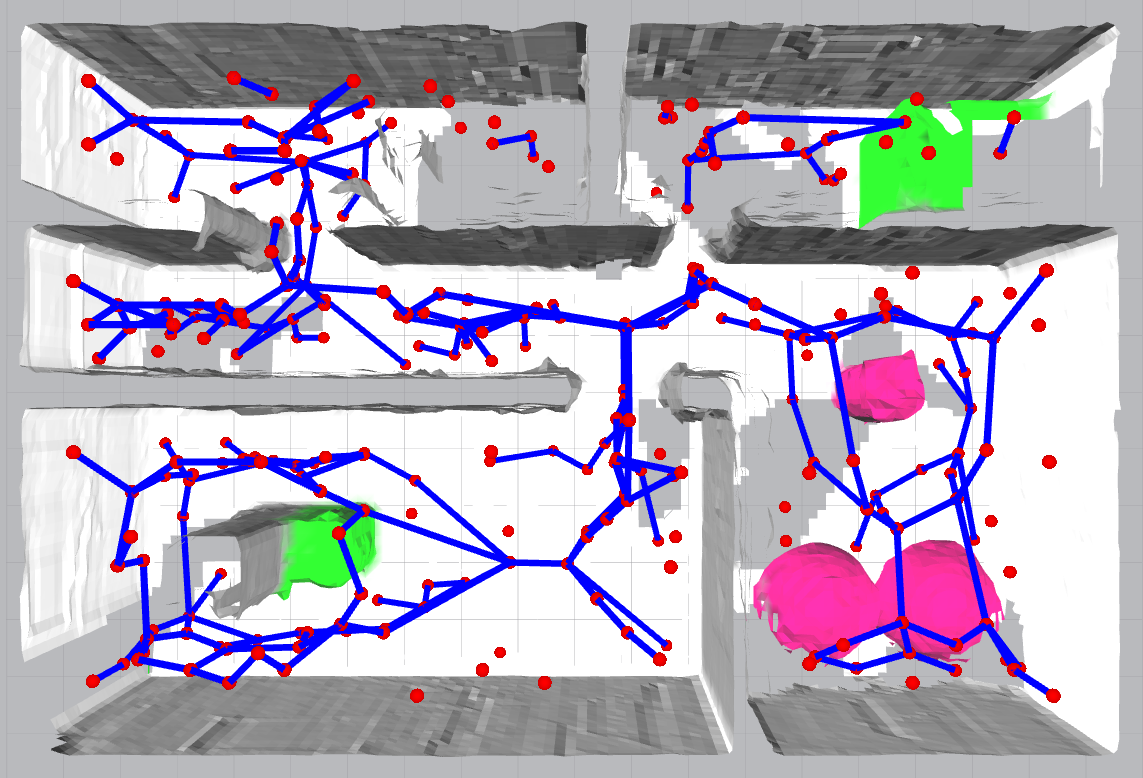}&
    \includegraphics[width=.30\columnwidth,trim=000 000 0 0 mm, clip=true]{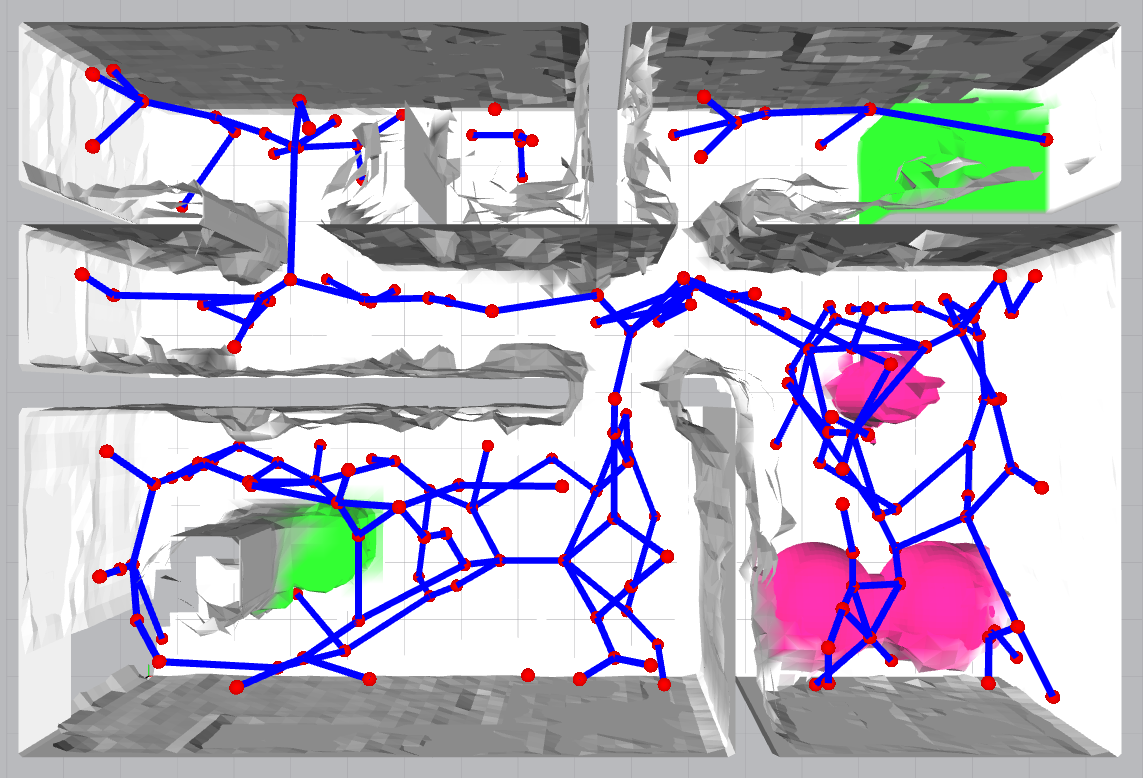}\\
    \rowname{$\sigma = 0.1$}&
    \includegraphics[width=.30\columnwidth,trim=000 000 0 0 mm, clip=true]{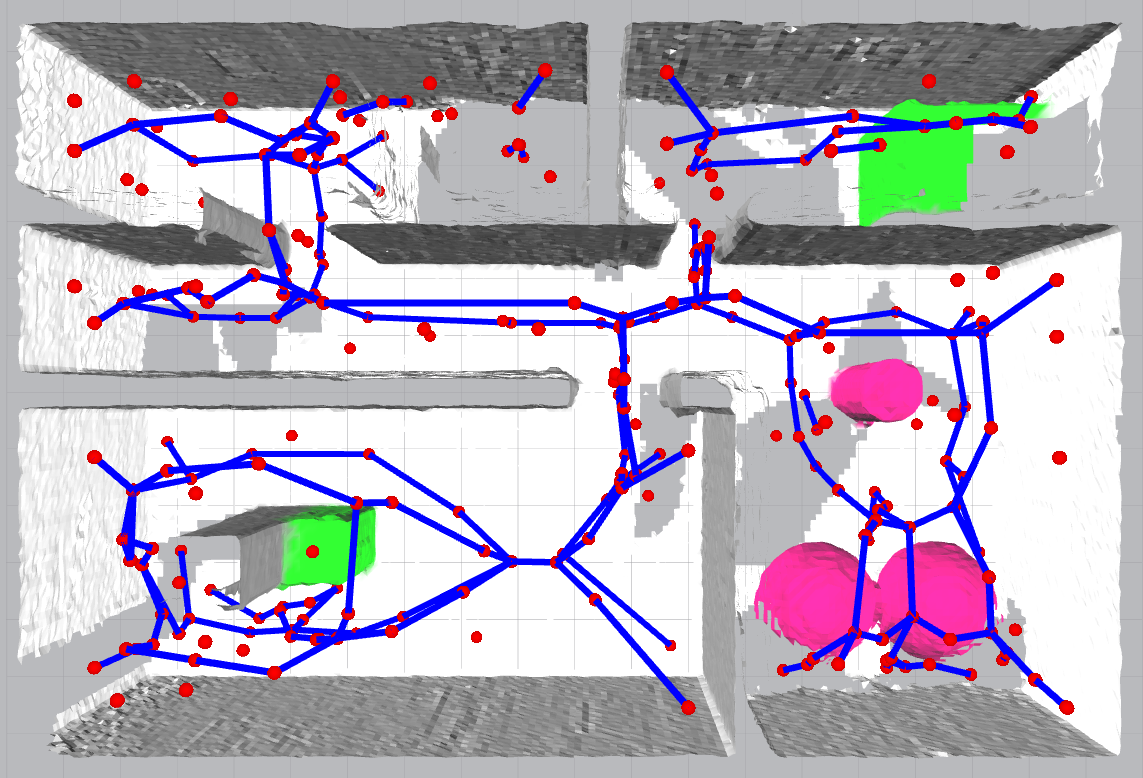}&
    \includegraphics[width=.30\columnwidth,trim=000 000 0 0 mm, clip=true]{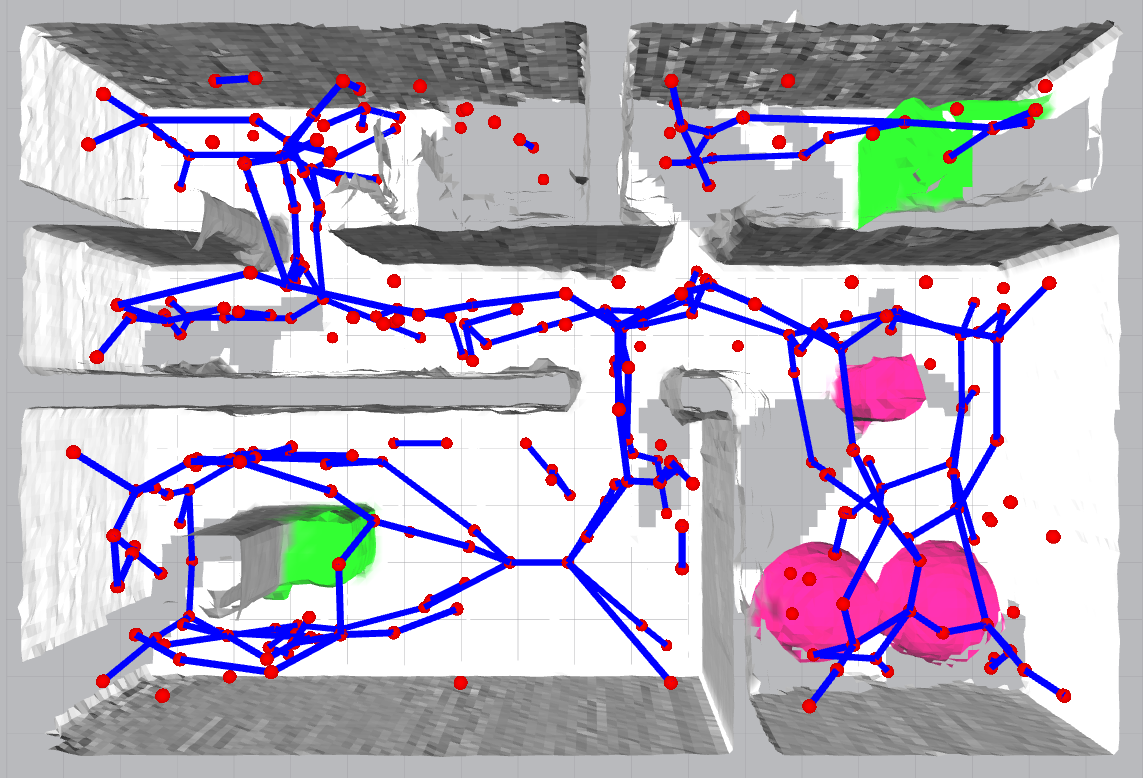}&
    \includegraphics[width=.30\columnwidth,trim=000 000 0 0 mm, clip=true]{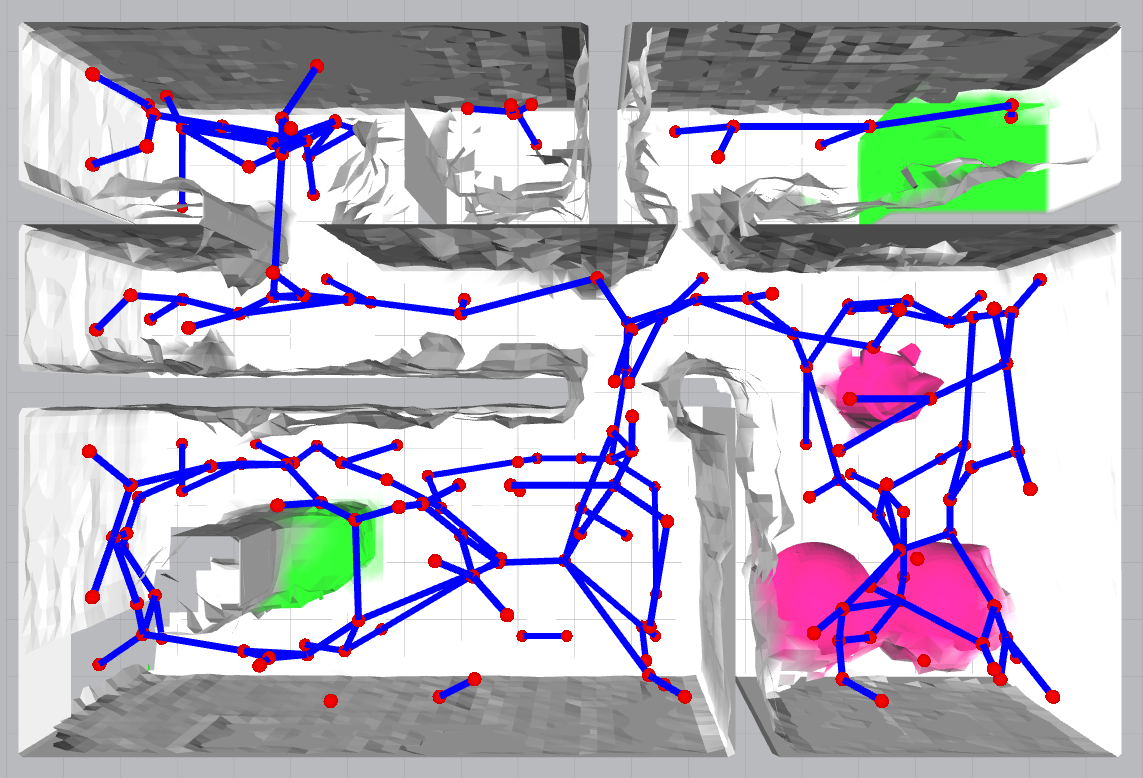}\\
    \rowname{$\sigma = 0.2$}&
    \includegraphics[width=.30\columnwidth,trim=000 000 0 0 mm, clip=true]{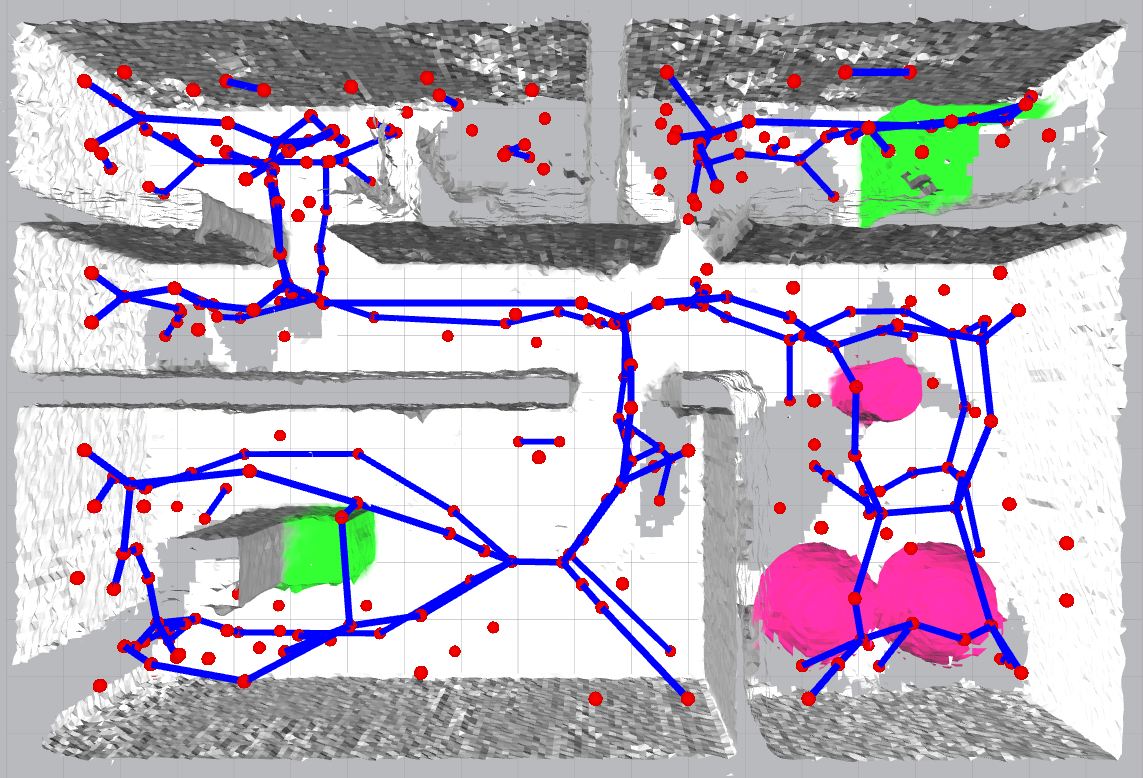}&
    \includegraphics[width=.30\columnwidth,trim=000 000 0 0 mm, clip=true]{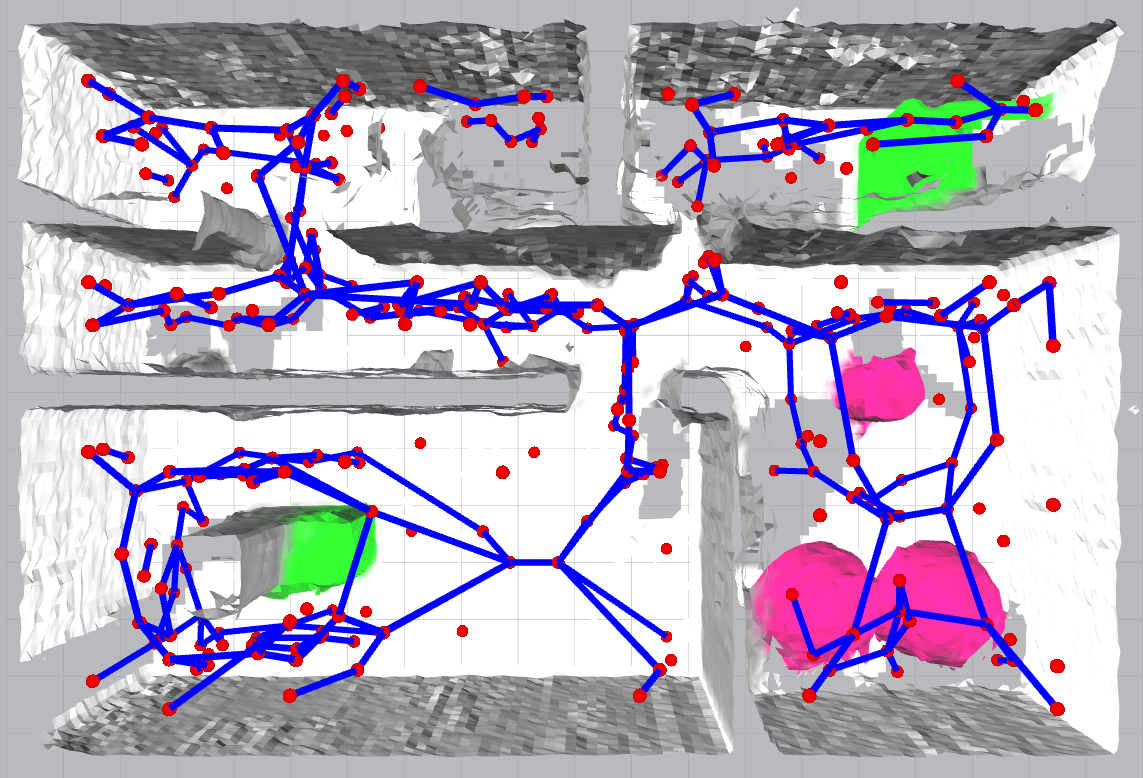}&
    \includegraphics[width=.30\columnwidth,trim=000 000 0 0 mm, clip=true]{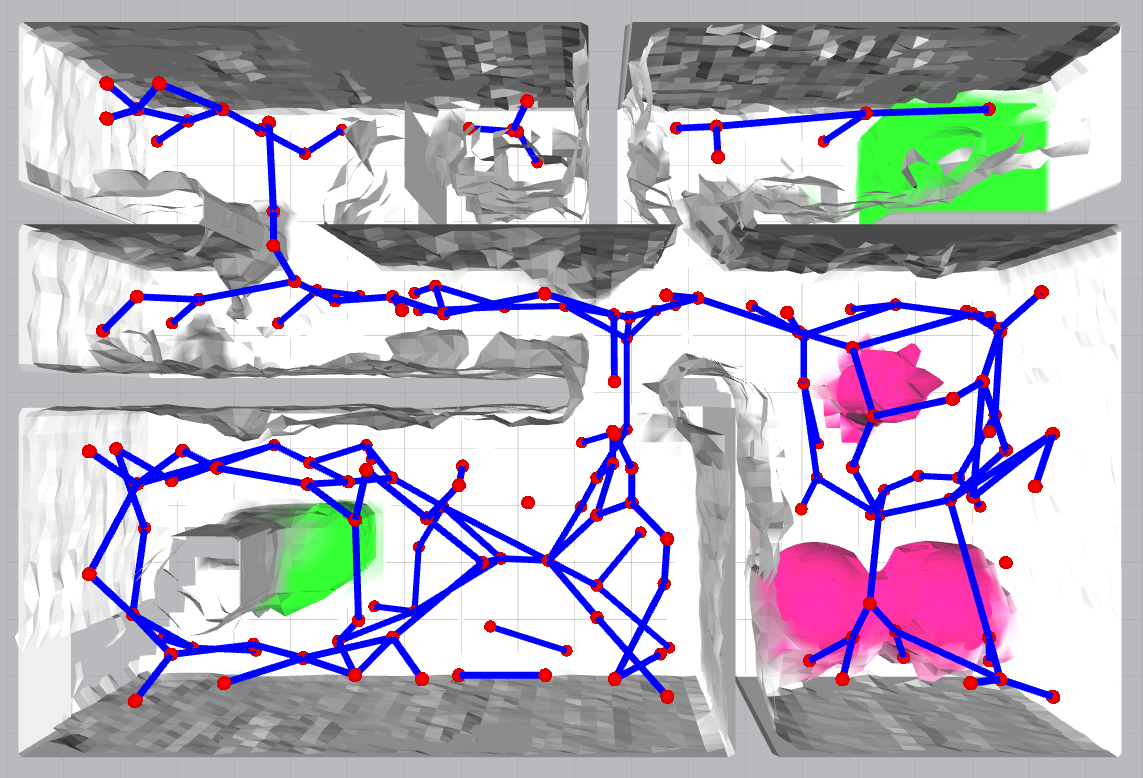}\\
    &0.10 m & 0.15 m & 0.25 m 
  \end{tabular}
  \caption{A comparison of sparse graphs generated from the same simulated environment, under different voxel sizes and amounts of noise in the distance measurements. The top row is simulated from ground truth ESDF data, while the other 3 rows show maps created from 200 simulated robot poses with a noisy depth sensor. As can be seen, despite extraneous edges introduced by noise or discretization error, the core structure of the graph remains the same. The diagram was generated with a minimum distance of $0.4$ meters.}
  \label{fig:graph_compare}
\end{figure}
%%trim option's parameter order: left bottom right top

\begin{figure}[tb]
  \centering
  \includegraphics[width=0.8\columnwidth,trim=0 0 0 0 mm, clip=true]{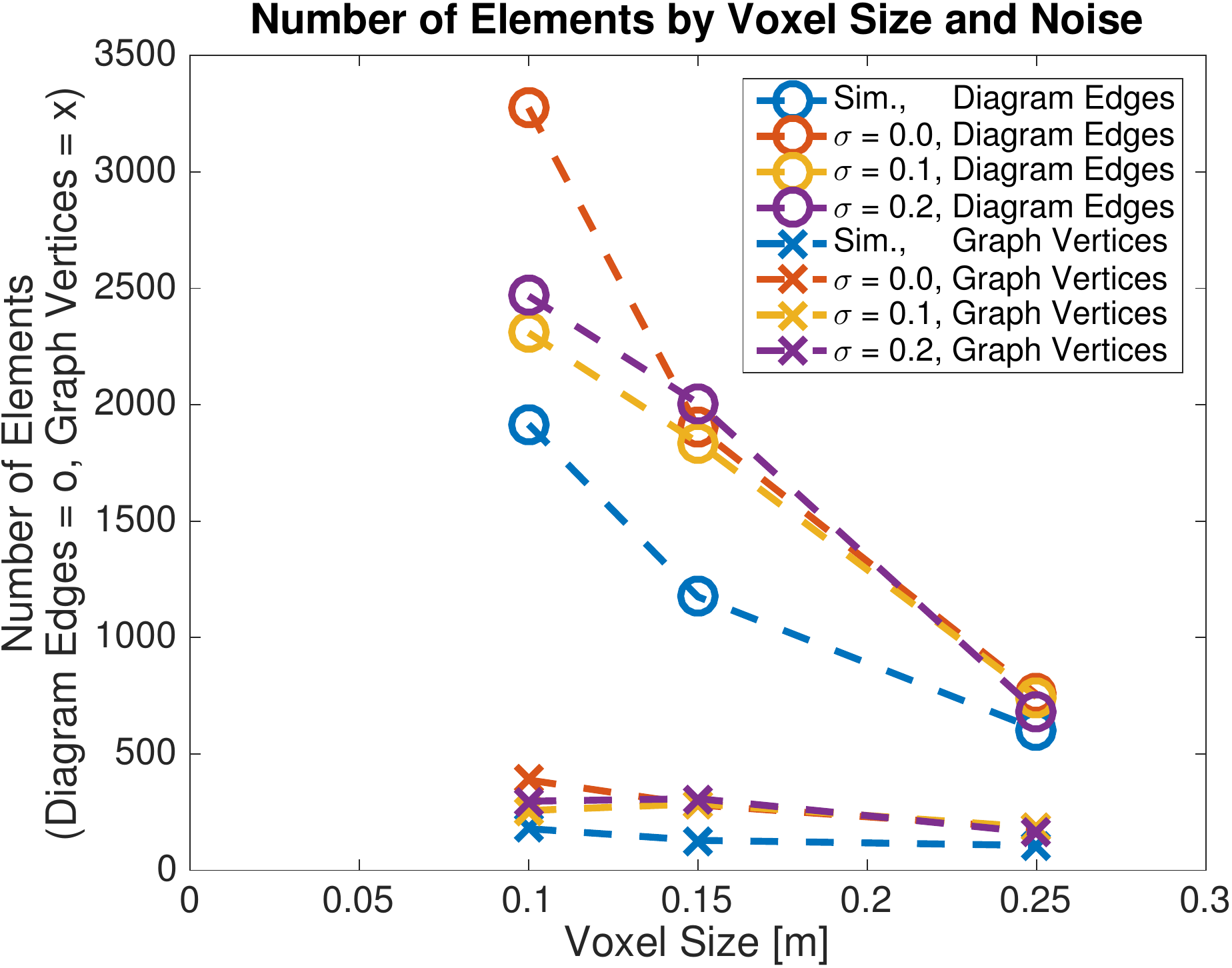}
  \caption{A comparison of the number of elements in the skeleton diagram (circles) and sparse graph (crosses). As can be seen, the voxel size and, to a lesser extent, noise level have a strong impact on the number of elements in the diagram, but the graph size stays mostly the same. This implies that the graph size is a function of the topology of the space, not the resolution or quality of the map.}
  \label{fig:sim_graph_results}
\end{figure}
%trim option's parameter order: left bottom right top

We aim to show that the sparse graph construction method preserves the underlying structure of the scene, even in the presence of noise or different levels of discretization error (different voxel sizes).
To do so, we construct a simulation world for which we have both the ground truth ESDF and are able to simulate synthetic viewpoints within this space, optionally corrupted by noise.

We generate the skeleton diagram and sparse graph for a variety of voxel sizes and noise magnitudes, shown in \reffig{fig:graph_compare}.
The top row shows the sparse graph built from perfect ground truth data acquired from the simulation.
The following nine scenarios show the same simulated world, reconstructed from 200 randomly selected robot poses with a simulated depth sensor. The sensor has a $90\deg$ field of view and $320\times240$ resolution.
Optionally, we apply independent Gaussian noise to each distance measurement, with $\sigma = 0.1$ or $0.2$.
These scans are then incrementally integrated into a Truncated Signed Distance Field (TSDF), and we construct the ESDF from this TSDF using full Euclidean distances, as described in our previous work~\cite{oleynikova2017voxblox}.

While it can be seen that noise and discretization error corrupt the graph by adding extraneous edges, the fundamental structure of the graph remains the same. 

Another important effect of the sparse graph representation is that its size stays largely constant regardless of the size of the underlying diagram.
This is shown in \reffig{fig:sim_graph_results}, where we compare the number of elements on the skeleton diagram (circles) and in the sparse graph (crosses).
Even though the voxel size (and to a lesser extent, noise level) has a large impact on the number of elements in the skeleton diagram, the number of elements in the graph remains almost constant. 
This shows the robustness of our approach to different resolutions and corruption by noise.

\subsection{Planning}
\label{sec:planning_results}
\begin{figure}[tb]
  \centering
  \includegraphics[width=0.8\columnwidth,trim=0 0 0 0 mm, clip=true]{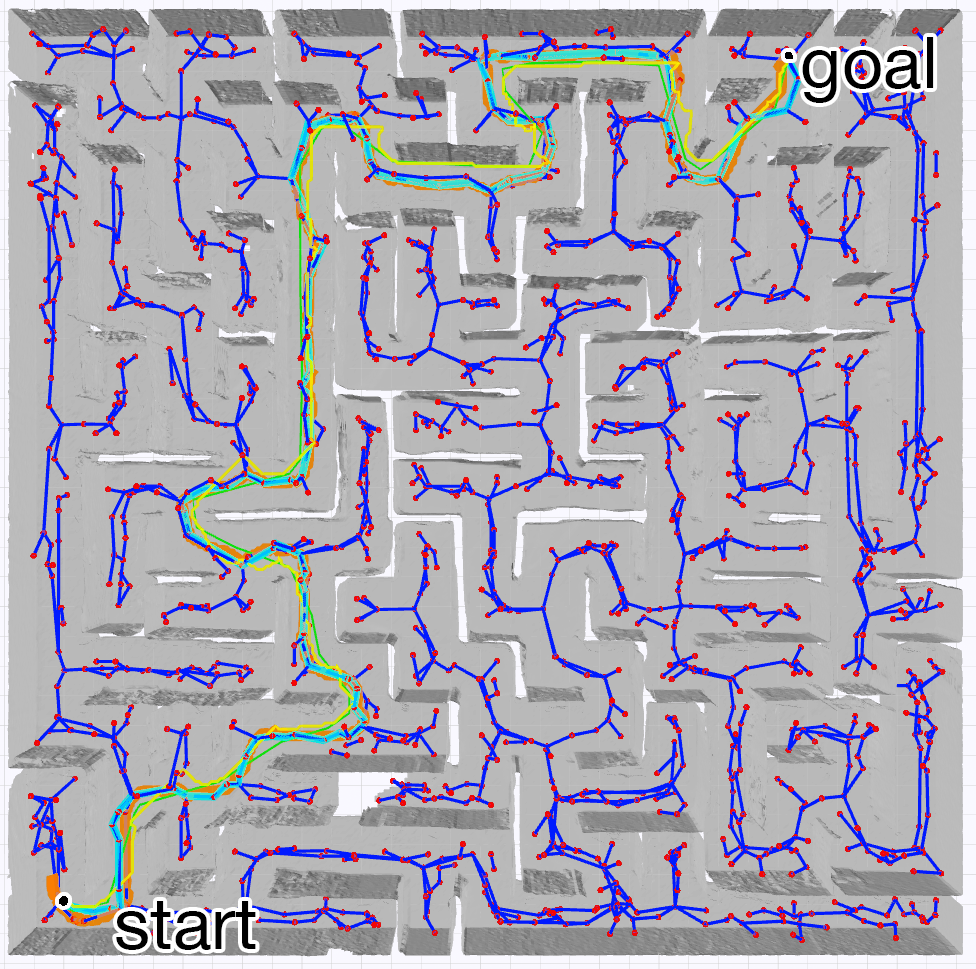}
  \caption{Sparse graph (dark blue edges and red vertices) and planning results through a 30 m by 30 m maze, autonomously explored by an MAV in simulation. Green shows the first RRT* path to the goal, yellow is A* through the ESDF, orange is A* through the skeleton diagram, and cyan is the search through the sparse graph.}
  \label{fig:maze_results}
\end{figure}
%trim option's parameter order: left bottom right top

\begin{table}[tb]
  \centering
  \begin{tabular}{lrrr}
    \toprule
    \tableheader{1.5cm}{\textbf{Planner}} & \tableheader{1.1cm}{\textbf{Time} [$s$]} & \tableheader{1.6cm}{\textbf{Path Length} [$m$]} &  \tableheader{1.1cm}{\textbf{Solution Vertices}} \\
    \midrule[1.5pt]
    RRT* & 2.500 & \textbf{62.079} & \textbf{39} \\
    RRT Connect & 0.1530 & 108.07 & 68 \\
    A* ESDF & 452.3 & 72.26 & 627  \\
    A* Skeleton Diagram & 0.0503 & 92.876 & 732 \\
    Sparse Graph & \textbf{0.00328} & 86.178 & 110 \\
    \bottomrule
  \end{tabular}
  \caption{Quantitative results for planning through the maze, shown in \reffig{fig:maze_results}. Our method (sparse graph) is 800 times faster than the initial solution of RRT*, and even 50 times faster than the RRT Connect. It produces slightly longer paths since it is bound to the maximum-clearance graph.}
  \label{table:maze_table}
\end{table}

To evaluate the ability of our method to very quickly generate initial feasible paths through complex environments, we set up a maze environment in simulation, and allowed a simulated MAV equipped with a forward-facing stereo camera to autonomously explore it using a next-best-view exploration algorithm~\cite{bircher2016receding}.
The maze environment is 30 m by 30 m, represented by a map with 10 cm voxels.
We then take the map created from this exploration and generate the ESDF, skeleton diagram, and sparse graph from it.

We use the methods described in \refsec{sec:planners} to plan between two locations in the maze, shown in \reffig{fig:maze_results}.
As can be seen, the sparse graph representation of the maze (red vertices and dark blue edges) appears to be a good descriptor of the underlying structure.
While all the planning methods find largely similar paths, there are a few differences, since ESDF A* (yellow) and RRT* (green) are not bound to staying on the diagram (orange) or sparse graph (cyan).

However, the major difference is in the execution time, shown in \reftab{table:maze_table}.
The RRT* takes approximately 2.5 seconds to find the initial solution, while an A* search through the complete ESDF takes over 7 minutes.
In contrast, a search through the diagram takes only 50 ms, and the sparse graph search takes only 3 ms.

It is also interesting to look at the path lengths; clearly our proposed method does not produce the shortest paths.
It does produce the maximum-clearance path through this space, and can be further shortened by polynomial smoothing and optimization, as described in \refsec{sec:trajectory}.
The most important advantage of our method is that it takes almost 3 orders of magnitude less time to find this initial path using the sparse graph search than using RRT*.

%\begin{easylist}[itemize]
%  & Map compression
%  && $\%$ size compression
%  && Reconstruction accuracy based on strength of pruning
%  & Free space coverage
%  && What percentage of the free space is covered in various scenarios?
%  & Robustness/consistency
%  && Generate multiple maps of same area, corrupted by noise
%  && How different are the skeletons generated?
%  & Different voxel size
%  && How similar are the graphs?
%  & Compare to RRT* + smoothing
%  && For same runtimes, compare results
%  && For ``best" RRT* solution, compare results vs. fixed runtime of ours
%  & Compare to just doing straight-up A* in the 26-conn edge graph
%  && Ok this is probably fast, but like... advantages/disadvantages?
%  & $\theta$-SMA versus how noise-tolerant the graph is
%  && Increase noise in simulated scenario, see how much the GVD changes as $\theta$ changes as well
%  && See how much the resulting sparse graph changes as $\theta$ changes
%  & Graph sparsification: how many voxels in layer vs. how many elements in graph
%  && Also edge voxels vs. sparse edges, vertex voxels vs. sparse vertices (with non-max supression)
%\end{easylist}

\subsection{Platform Experiments}
\begin{figure}[tb]
  \centering
  \includegraphics[width=1.0\columnwidth,trim=183 0 135 70 mm, clip=true]{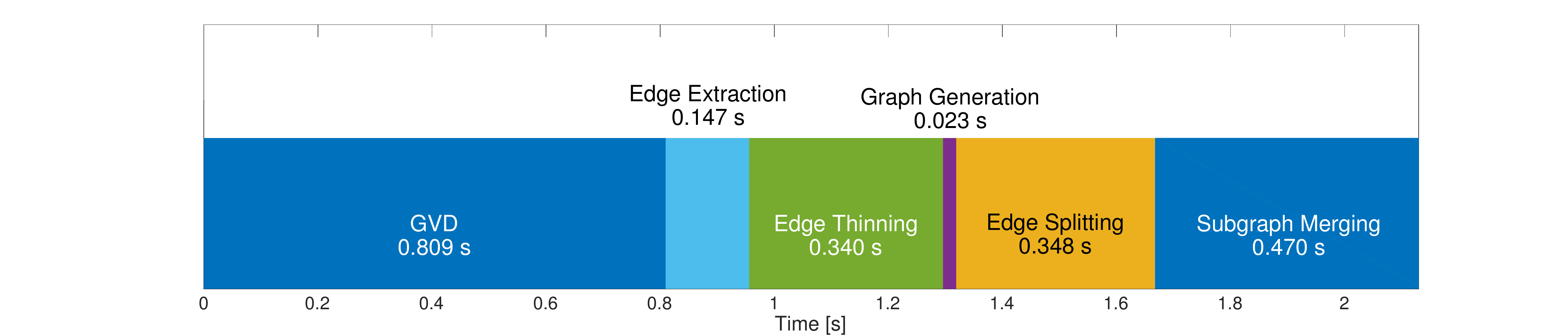}
  \caption{Timing information for the platform experiments, shown in \reffig{fig:teaser}. The map is 20 m by 22 m at 0.07 m voxel resolution. The total time to generate the sparse graph from the ESDF was 2.13 seconds on the MAV on-board computer.}
  \label{fig:timings}
\end{figure}
%trim option's parameter order: left bottom right top
To demonstrate the feasibility of this method on real data, we performed a flight through a machine hall with a custom-built drone, using ORB SLAM2 for state estimation and loop closure, and dense mapping of RGB-D data through the submapping approach described by Millane \etal~\cite{millane2017tsdf}.
This shows the intended use-case of this work: performing an initial exploratory flight with an MAV, quickly constructing a skeleton of the optimized map as shown in \reffig{fig:timings}, and then using the topological map to rapidly plan return paths.

The results are shown in \reffig{fig:teaser}, where the sparse graph edges are shown in black, an initial path through the graph in yellow, and we use the techniques described in our previous work to get an initial dynamically-feasible polynomial path through a subset of the vertices (orange) and optimize it to be collision-free (red)~\cite{oleynikova2016continuous-time,oleynikova2017safe}.
A video showing the experiments and results is available at \href{https://youtu.be/U_6rk-SF0Nw}{\texttt{youtu.be/U\_6rk-SF0Nw}}.

This shows that we are able to quickly produce useful topological graphs to aid in global path planning from real sensor data from an MAV.

%\begin{easylist}
%  & Show sparse topology on real noisy map
%  & Plan in this space
%  & Show that a dynamically-feasible smooth path is produced (how?)
%  & Timing results?
%\end{easylist}

%Platform experiments
%\todo{???????}
%\begin{easylist}[itemize]
%  & Show on platform
%  & Probably in the office, let's be honest
%  & Or some other cool indoor environment
%\end{easylist}

\section{Conclusions}
In this paper, we proposed a method to build 3D skeleton diagrams and sparse graphs that maintain the topology and connectivity of the original space, by using a combination of techniques from 2D Voronoi Diagram-based planning and 3D graphics skeletonization literature.
Our method is robust to noise and resolution changes, and is shown to speed up global planning search queries by up to 800 times over initial solutions to RRT*.
We also show its applicability to real maps built in-flight on an MAV, and demonstrate how our graph can be used to plan optimized trajectories based on the fast initial solution.
Future work would include adding clearance information to the sparse graph and making the method completely incremental.

\bibliographystyle{ieeetr}

\bibliography{sources}

\end{document}